\documentclass[twocolumn,a4paper,draftcls,14pt]{IEEEtran}
\usepackage{amsmath} 
\usepackage{amssymb}
\usepackage{graphicx}
\usepackage{subfigure} 
\usepackage{cite}
\usepackage{url}
\usepackage[font=normalsize,labelfont=bf]{caption}
\urldef{\mailsa}\path|{dev.kakde, arin.chaudhuri,  seunghyun.kong, maria.jahja|
\urldef{\mailsb}\path|hansi.jiang,jorge.silva, anya.mcguirk}@sas.com|

\begin{document} 
\title{Peak Criterion for Kernel Bandwidth Selection for Support Vector Data Description}
\author{\
\IEEEauthorblockN{
Deovrat Kakde\IEEEauthorrefmark{1},
Arin Chaudhuri\IEEEauthorrefmark{2}, 
Seunghyun Kong\IEEEauthorrefmark{3},
Maria Jahja\IEEEauthorrefmark{4},
Hansi Jiang\IEEEauthorrefmark{5},
Jorge Silva\IEEEauthorrefmark{6} and
Anya Mcguirk\IEEEauthorrefmark{5},
}

\IEEEauthorblockA{
Advanced Analytics Divison,
SAS Institute\\
Cary, NC, USA\\
Email: \IEEEauthorrefmark{1}Dev.Kakde@sas.com,
\IEEEauthorrefmark{2}Arin.Chaudhuri@sas.com,
\IEEEauthorrefmark{3}Seunghyun.Kong@sas.com,
\IEEEauthorrefmark{4}Maria.Jahja@sas.com,
\IEEEauthorrefmark{5}Hansi.Jiang@sas.com,
\IEEEauthorrefmark{6}Jorge.Silva@sas.com,
\IEEEauthorrefmark{7}Anya.Mcguirk@sas.com,
}
}

\maketitle

\begin{abstract} 
Support Vector Data Description (SVDD) is a machine-learning technique
used for single class classification and outlier detection. SVDD
formulation with kernel function provides a flexible boundary around
data. The value of kernel function parameters affects the nature of the data
boundary. For example, it is observed that with a Gaussian kernel, as the
value of kernel bandwidth is lowered, the data boundary changes from
spherical to wiggly. The spherical data boundary leads to underfitting,
and an extremely wiggly data boundary leads to overfitting. In this paper,
we propose an empirical criterion to obtain good values of the Gaussian
kernel bandwidth parameter. This criterion provides a smooth boundary that captures the 
essential geometric features of the data.
\end{abstract} 

\section{Introduction}
\label{intro}
Support Vector Data Description (SVDD) is a machine learning technique
used for single-class classification and outlier detection. SVDD
is similar to Support Vector Machines and was first introduced
by Tax and Duin \cite{tax2004support}. It can be used to build a flexible boundary around
single-class data. The data boundary is characterized by observations
designated as support vectors.
SVDD is used in domains where the majority of data belongs to a single
class. Several researchers have proposed use of SVDD for multivariate process
control \cite{sukchotrat2009one, camci2008general}. Other applications of SVDD involve machine condition monitoring \cite{widodo2007support, ypma1999robust} and image classification \cite{sanchez2007one}.
\subsection{Mathematical Formulation}
{\bf Normal Data Description:}\\
The SVDD model for normal data description builds a minimum radius hypersphere around the data.\\ 
\\
{\bf Primal Form:}\\
Objective Function:
\begin{equation}
\min R^{2} + C\sum_{i=1}^{n}\xi _{i}, 
\end{equation}
subject to: 
\begin{align}
 \|x _{i}-a\|^2 \leq R^{2} + \xi_{i}, \forall i=1,\dots,n,\\
 \xi _{i}\geq 0, \forall i=1,...n.
\end{align}
where:\\
$x_{i} \in {\mathbb{R}}^{m}, i=1,\dots,n  $ represents the training data,\\
$ R:$ radius, represents the decision variable,\\
$\xi_{i}:$ is the slack for each variable,\\
$a$: is the center, a decision variable, \\
$C=\frac{1}{nf}:$ is the penalty constant that controls the trade-off between the volume and the errors, and,\\
$f:$ is the expected outlier fraction.\\ \ \\
{\bf Dual Form:}\\
The dual formulation is obtained using the Lagrange multipliers.\\ 
Objective Function:
\begin{equation} 
    \max\ \sum_{i=1}^{n}\alpha _{i}(x_{i}.x_{i}) - \sum_{i,j}^{ }\alpha _{i}\alpha _{j}(x_{i}.x_{j}) ,
\end{equation}
subject to:
\begin{align}
& &  \sum_{i=1}^{n}\alpha _{i}  = 1,\\
 & & 0 \leq  \alpha_{i}\leq C,\forall i=1,\dots,n.
\end{align}
where:\\
$\alpha_{i}\in \mathbb{R}$: are the Lagrange constants,\\
$C=\frac{1}{nf}:$ is the penalty constant.\\ \ \\
{\bf Duality Information:}\\
Depending upon the position of the observation, the following results hold good:

Center Position: \begin{equation} \sum_{i=1}^{n}\alpha _{i}x_{i}=a. \end{equation}
Inside Position: \begin{equation} \left \| x_{i}-a \right \| < R \rightarrow \alpha _{i}=0.\end{equation}
Boundary Position: \begin{equation} \left \| x_{i}-a \right \| = R \rightarrow 0< \alpha _{i}< C.\end{equation}
Outside Position: \begin{equation}\left \| x_{i}-a \right \| > R \rightarrow \alpha _{i}= C.\end{equation}
The radius of the hypersphere is calculated as follows:\\
\begin{equation}   
R^{2}=(x_{k}.x_{k})-2\sum_{i}^{ }\alpha _{i}(x_{i}.x_{k})+\sum_{i,j}^{ }\alpha _{i}\alpha _{j}(x_{i}.x_{j}).
 \end{equation}
$ \forall x_{k} \in SV_{<C} $
, where $SV_{<C}$  is the set of support vectors that have $ \alpha _{k} < C $.

{\bf Scoring:}

For each observation $ z $  in the scoring data set, the distance $ \operatorname{dist}^{2}(z) $ is calculated as follows: 
\begin{equation}  \operatorname{dist}^{2}(z)=(z.z) - 2\sum_{i}^{ }\alpha _{i}(x_{i}.z) +\sum_{i}^{ }\alpha_{i,j}\alpha _{j}(x_{i}.x_{j}). \end{equation}
The scoring data set points with $ \operatorname{dist}^{2}(z) > R^{2} $ are designated as outliers.

The circular data boundary can include a significant amount of space with a very sparse distribution of training observations. Scoring with this model can lead to many outliers being classified as in-liers.
 Hence, instead of a circular shape, a compact bounded outline around the data is often desired. Such an outline should approximate the shape of the single-class training data. This is possible with the use of kernel functions.\\
 
{\bf Flexible Data Description:}

The Support Vector Data Description is made flexible by replacing the inner product $ (x_{i}.x_{j}) $ with a suitable kernel function $ K(x_{i},x_{j}) $. The Gaussian kernel function used in this paper is defined as:
\begin{equation}  
K(x_{i}, x_{j})= \exp  \dfrac{ -\|x_i - x_j\|^2}{2s^2}
 \end{equation}
where $s$ is the Gaussian bandwidth parameter.

The modified mathematical formulation of SVDD with kernel function is as follows:

Objective function:
\begin{equation}  \label{dualob}\
   \max\ \sum_{i=1}^{n}\alpha _{i}K(x_{i},x_{i}) - \sum_{i,j}^{ }\alpha _{i}\alpha _{j}K(x_{i},x_{j}),
\end{equation}\
subject to:
\begin{align}
 & &\sum_{i=1}^{n}\alpha _{i} = 1,\\
 & & 0 \leq  \alpha_{i}\leq C , \forall i=1,\dots,n.
\end{align}
The results (7) through (10) hold good when the kernel function is used in the mathematical formulation.\\
The threshold $R^{2}$ is calculated as :
\begin{multline}
R^{2} = K(x_{k},x_{k})-2\sum_{i}^{ }\alpha _{i}K(x_{i},x_{k})\\+\sum_{i,j}^{ }\alpha _{i}\alpha _{j}K(x_{i},x_{j})
\end{multline}
$ \forall x_{k} \in SV_{<C} $, where $SV_{<C}$  is the set of support vectors that have $ \alpha _{k} < C $.

{\bf Scoring:}

For each observation $ z $  in the scoring data set, the distance $ \operatorname{dist}^{2}(z) $ is calculated as follows: 
\begin{multline}
 \operatorname{dist}^{2}(z)= K(z,z) - 2\sum_{i}^{ }\alpha _{i}K(x_{i},z)\\ +\sum_{i,j}^{ }\alpha _{i}\alpha _{j}K(x_{i},x_{j}).
\end{multline}
The scoring data set points with $\operatorname{dist}^{2}(z) > R^{2} $ are designated as outliers.

\subsection{Importance of Kernel Bandwidth Value}
The flexible data description is preferred when the data boundary
is non-spherical. The tightness of the boundary is a function of
the number of support vectors. In the case of a Gaussian kernel, it is observed
that if the value of outlier fraction $f$ is kept constant, the number of
support vectors identified by the SVDD algorithm is a function of the Gaussian
bandwidth parameter $s$. At very low values of $s$, the number of support vectors is high, approaching the number of observations. As the value of $s$
increases, the number of support vectors reduces. It is also observed
that at lower values of $s$, the data boundary is extremely wiggly.
As $s$ is increased, the data boundary becomes less wiggly,
and it starts to follow the general shape of the data. At higher values of $s$, the data boundary becomes more spherical. The selection of an
appropriate value of $s$ is tricky and often involves experimentation with
several values till a good data boundary is obtained. This paper
provides an empirical criterion for selecting a good value of the Gaussian
kernel bandwidth parameter. The corresponding data boundary is smooth
and captures essential visual features of the data.

The rest of the paper is organized as follows. Section~\ref{crit} illustrates
how data boundary changes with s using two-variable 
data sets of known geometry. The empirical criterion for selecting a good value of $s$ is
introduced and validated. Section~\ref{pd} provides
analysis of real-life data using the proposed method. Section~\ref{ss} details a simulation study conducted to evaluate the proposed method on random polygons.  A review of related work and comparison with existing methods are provided in Section~\ref{rw}.  Finally, conclusions and areas for further
research are provided in Section~\ref{cn}.

\section{Peak Criterion} 
\label{crit}

We experimented with several two-dimensional data sets of
known geometry to understand the relationship between data
boundary and choice of bandwidth parameter. We considered the data boundary to 
be of good quality if it closely follows the contours of the data shape.

As one might guess, the value of the objective function $ \left( \ref{dualob} \right) $  varies with the choice of bandwidth parameter, $ s $.  Denote this function:  $ V^{*}(s) $.  Our experimentation revealed that the optimal $ s $ seemed to occur at the first critical point(s) of the first derivative of $ V^{*} $ with respect to $ s $.  In other words, the best $ s $ occurred where the second derivative of $  V^{*}(s) $ equaled 0.  In the remainder of this paper, we explore the usefulness of choosing $ s $ utilizing these findings.  We refer to this method of selecting $ s $ as the Peak criterion. To examine the criterion's usefulness, we compute the first and second derivative values of $ V^{*}(s) $ with respect to $ s $ using the method of finite differences and thus, do not make any statements about the existence of analytical derivatives. 

To illustrate the approach and main findings of our experimentations, we focus on three data sets. These  data sets adequately illustrate and capture our general findings. The experimental approach and results are first explained in detail with a banana-shaped data set.  We then follow with the results obtained from a star-shaped data set and a data set with three non-overlapping data clusters.  

\begin{figure}[h]
    \centering
    \includegraphics[width=0.5\textwidth]{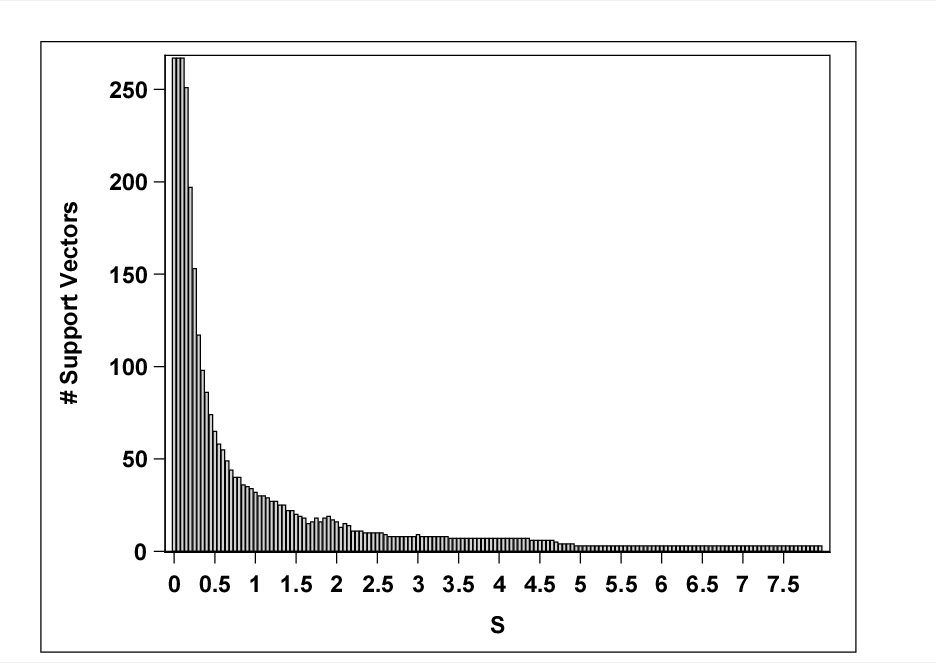}
    \caption{Number of support vectors vs. s: banana-shaped data}
    \label{fig:image_2}
\end{figure}

The two-dimensional banana-shaped data consists of 267 observations. The majority of the observations belong to a single
class, with very few outliers (fraction outliers, $f$=0.001). Figure \ref{fig:image_3}(a) provides a scatter plot of the data. 
To decide on a reasonable range of $ s $ to consider, we first examined how the number of support vectors varied with $s$ (see figure \ref{fig:image_2}). At low values of $ s $, a majority of the 267 observations are identified as support vectors. As $s$ increases, the number of support vectors generally decreases. For $s > 5$, the number of support vectors remains constant at 3. To cover all possible number of support vectors which can define the data boundary, we trained the data with the SVDD algorithm for $s$ in the interval $ [0.0001,8.0]$, in increments of 0.05, keeping $f$ constant at 0.001. 

At $s$ = 0.1, each point in the data is identified as a support vector, representing a very wiggly boundary
around the data. As the value of $s$ increases from 0.1 to 0.35, the data boundary is still wiggly, with many
``inside'' points identified as the support vectors. A very well defined boundary around the data is first observed at s=0.4. As s increases from 0.4 to 1.1, the boundary continues to conform to the Banana shape, with the number of support vectors decreasing from 86 to 30. Beyond $s$=1.1, the number of support vectors decreases and the boundary starts losing its true banana shape. For $s>=4$, the support vectors envelope the outer parabola of the Banana shape.
To confirm the shape of the data boundary, we score each training result on a 200x200 point data grid. Scoring results for select values of $s$ are provided in Figure \ref{fig:image_3}. 

\begin{figure}
\subfigure[Scatterplot of banana-shaped data]{\label{fig:image_1}\includegraphics[width=2.5in]{image1}}
\subfigure[s=0.2]{\label{fig:image_3}\includegraphics[width=2.5in]{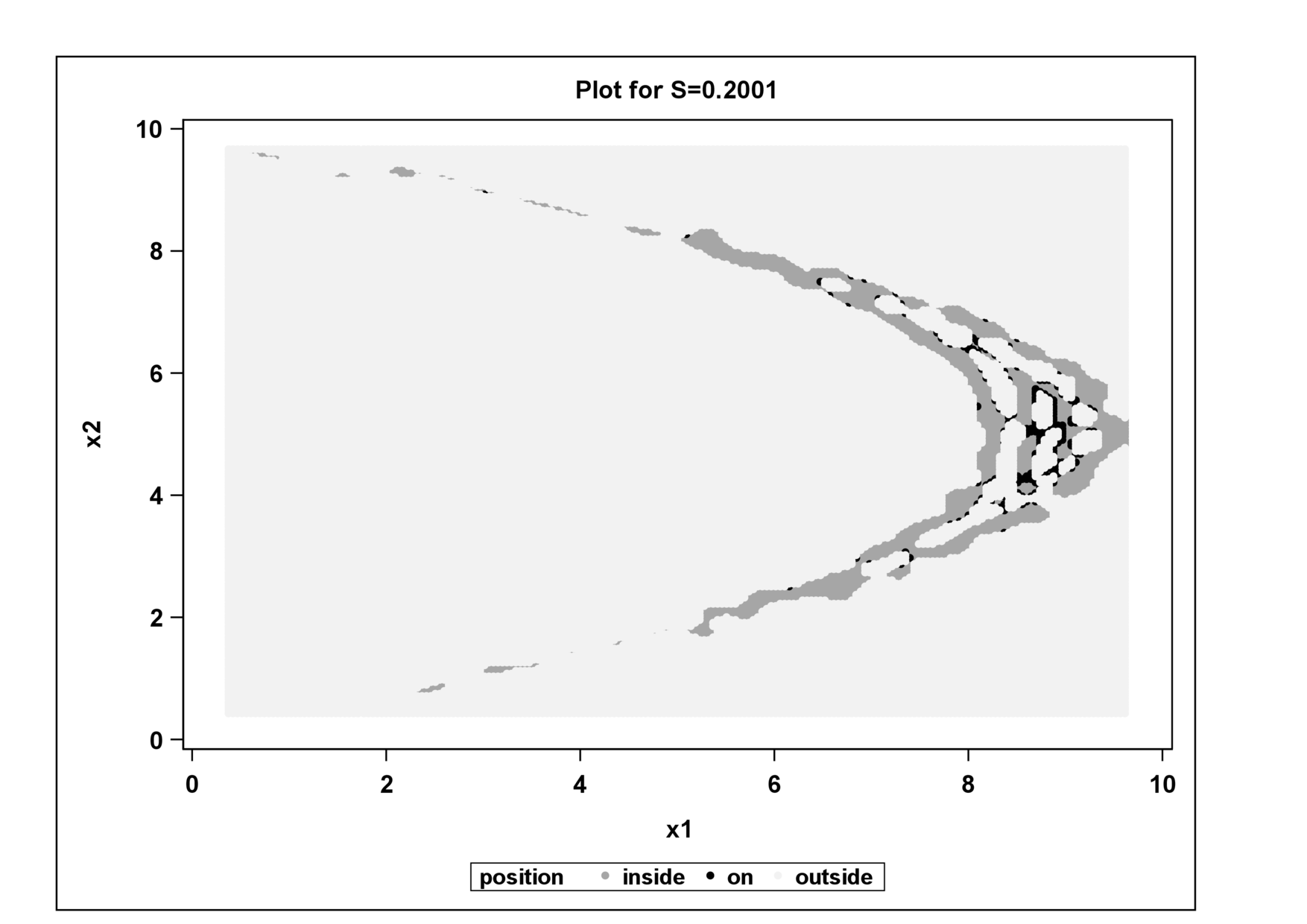}}
\subfigure[s=0.7]{\label{fig:image_4}\includegraphics[width=2.5in]{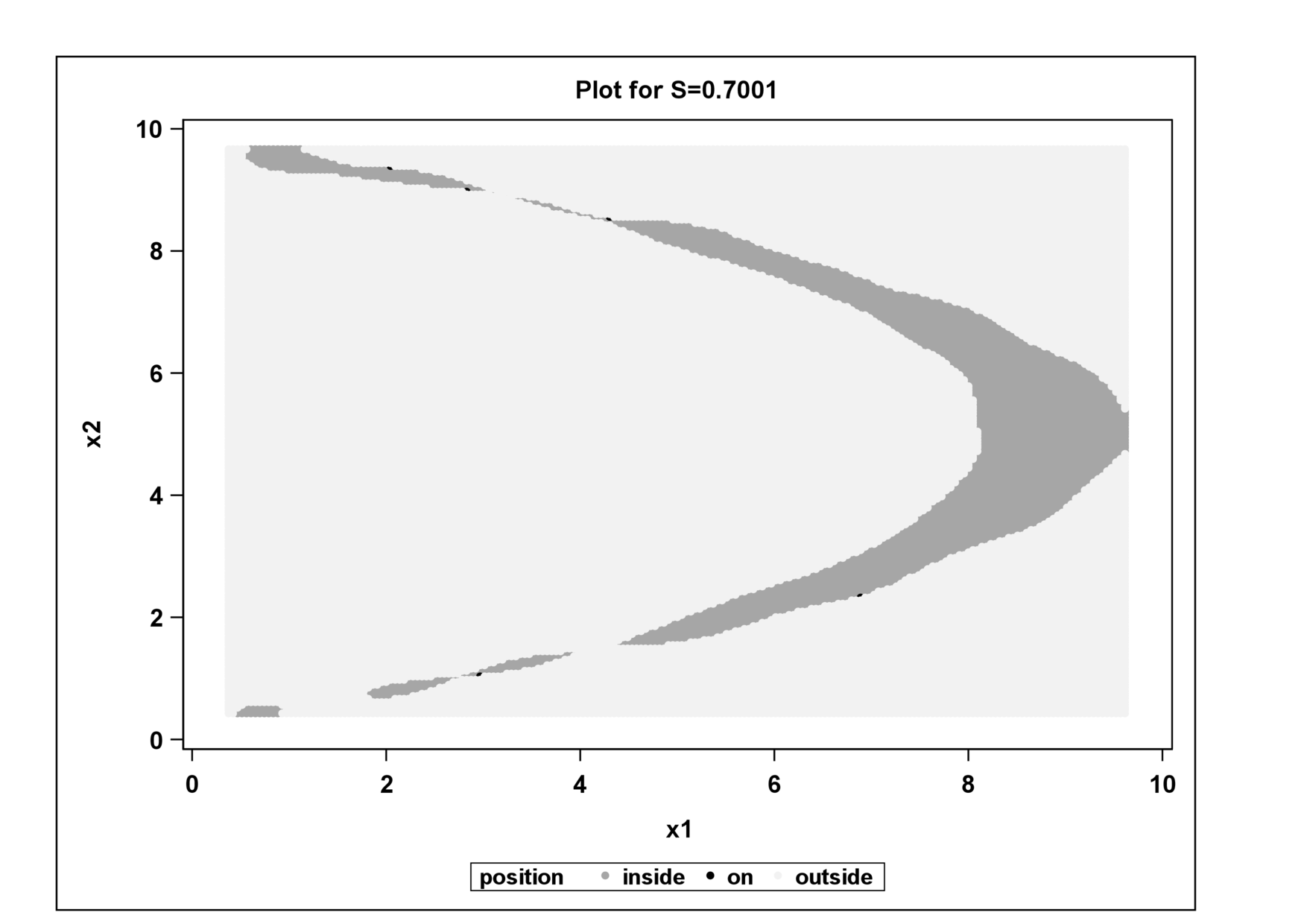}}
\subfigure[s=4.1]{\label{fig:image_5}\includegraphics[width=2.5in]{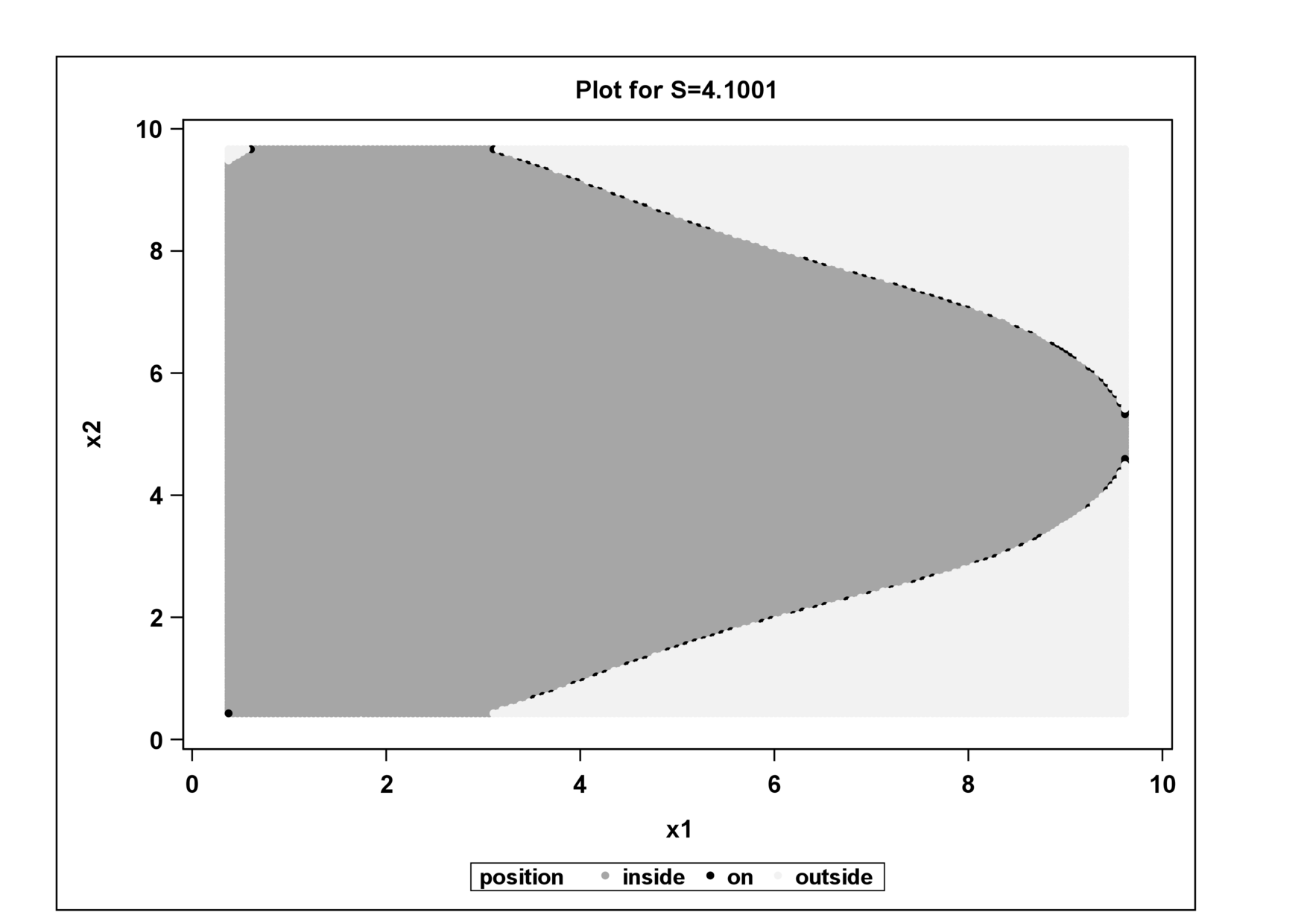}}
\caption{Data boundary for banana-shaped data. Fig (b) thru (d) show results of scoring on a 200x200 data grid. Light gray color indicates outside points, dark gray color indicates inside points and black color indicates support vectors.}\label{fig:image_3}
\end{figure}
\begin{figure}[h]
    \centering
    \includegraphics[width=0.5\textwidth]{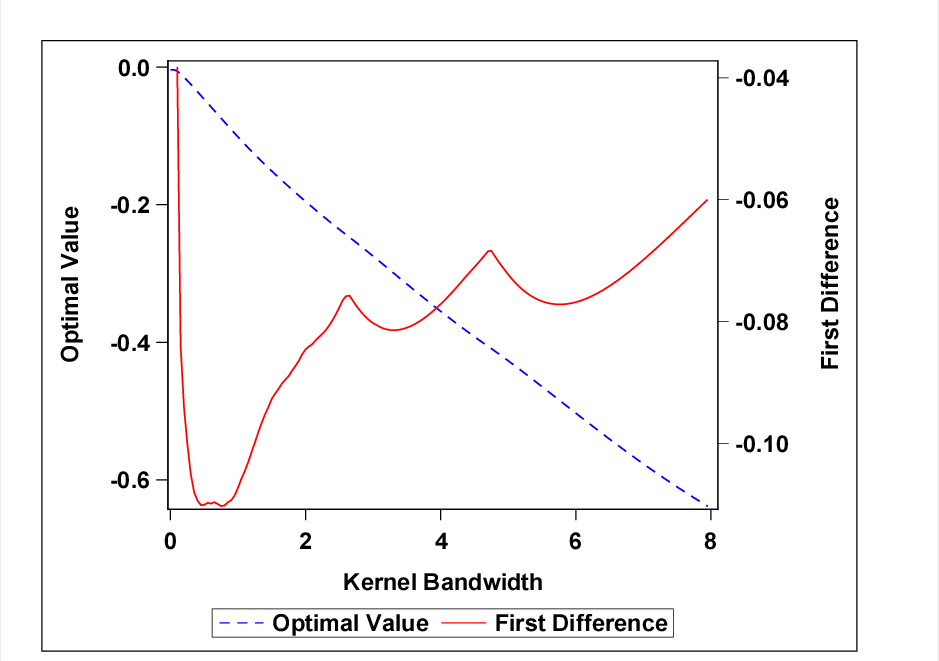}
    \caption{{Objective function value and first difference for banana-shaped data }}
    \label{fig:image_6}
\end{figure}

 Figure \ref{fig:image_6} shows  $ V^{*}(s) $, the value of dual objective function $ \left( \ref{dualob} \right) $ and its first derivative with respect to $s$, both plotted against $s$. $ V^{*}(s) $ is a decreasing function of $s$. As $s$ increases, the first derivative of $ V^{*}(s) $ first decreases. Between $s$=0.4 to $s$=1.1, it remains relatively flat indicating the derivative has reached its first critical point and that the optimal s occurs here.  After $s$=0.8, the first derivative starts to increase again. 

\begin{figure}[h]
 	\centering
 	\includegraphics[width=0.5\textwidth]{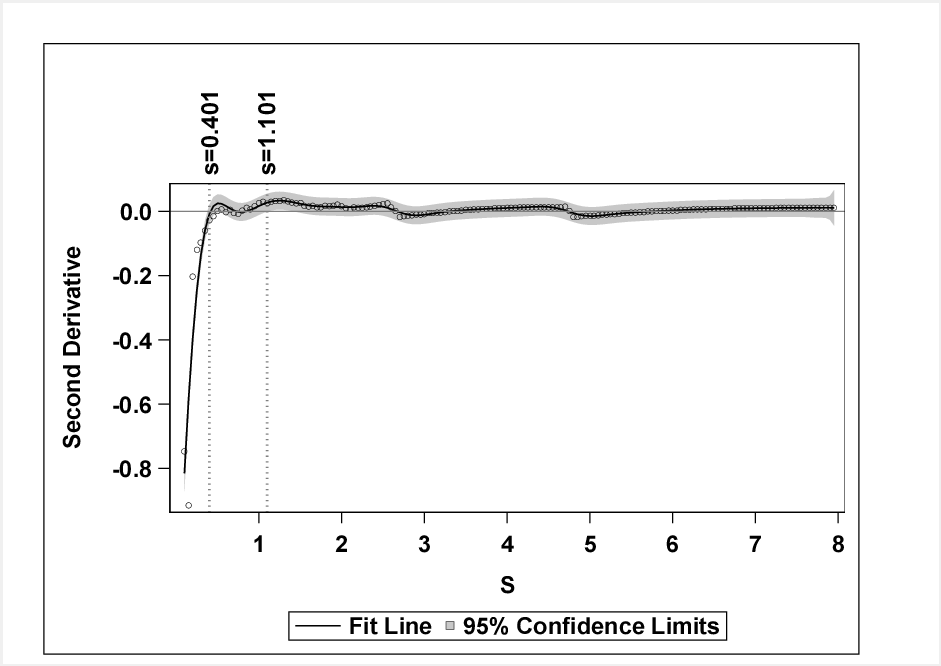}
 	\caption{Penalized B-spline fit for second derivative: banana-shaped data}
 	\label{fig:image_7}
 \end{figure}

Figure \ref{fig:image_7} shows the value of the second derivative of $ V^{*}(s) $, with respect to $s$ plotted against $s$. To decide if the value of the second derivative is zero, we fitted a penalized B-spline to the second derivative using the TRANSREG procedure available in the SAS software \cite{institute2015sas}. If the 95\% confidence interval of the fitted value of second derivative contains zero, we consider the second derivative value to be approximately zero.

As seen in Figure \ref{fig:image_7}, the second derivative is -0.20 at $s$=0.20. As $s$ increases, the value of the second derivative starts increasing. Between $s$=0.5 and 0.85, the second derivative is close to zero for the first time; we have the first set of first derivative critical points. All the values of $s$ in this range provide a data boundary of good quality. The data boundary using $s$=0.7 is shown in figure ~\ref{fig:image_3}(c). Compared to any other values of $s$ outside the range [0.4,1.1], this data boundary captures the essential geometric properties of the banana-shaped data.
 
\begin{figure}
\subfigure[Scatterplot of three-cluster data]{\label{fig:image9}\includegraphics[width=2.5in]{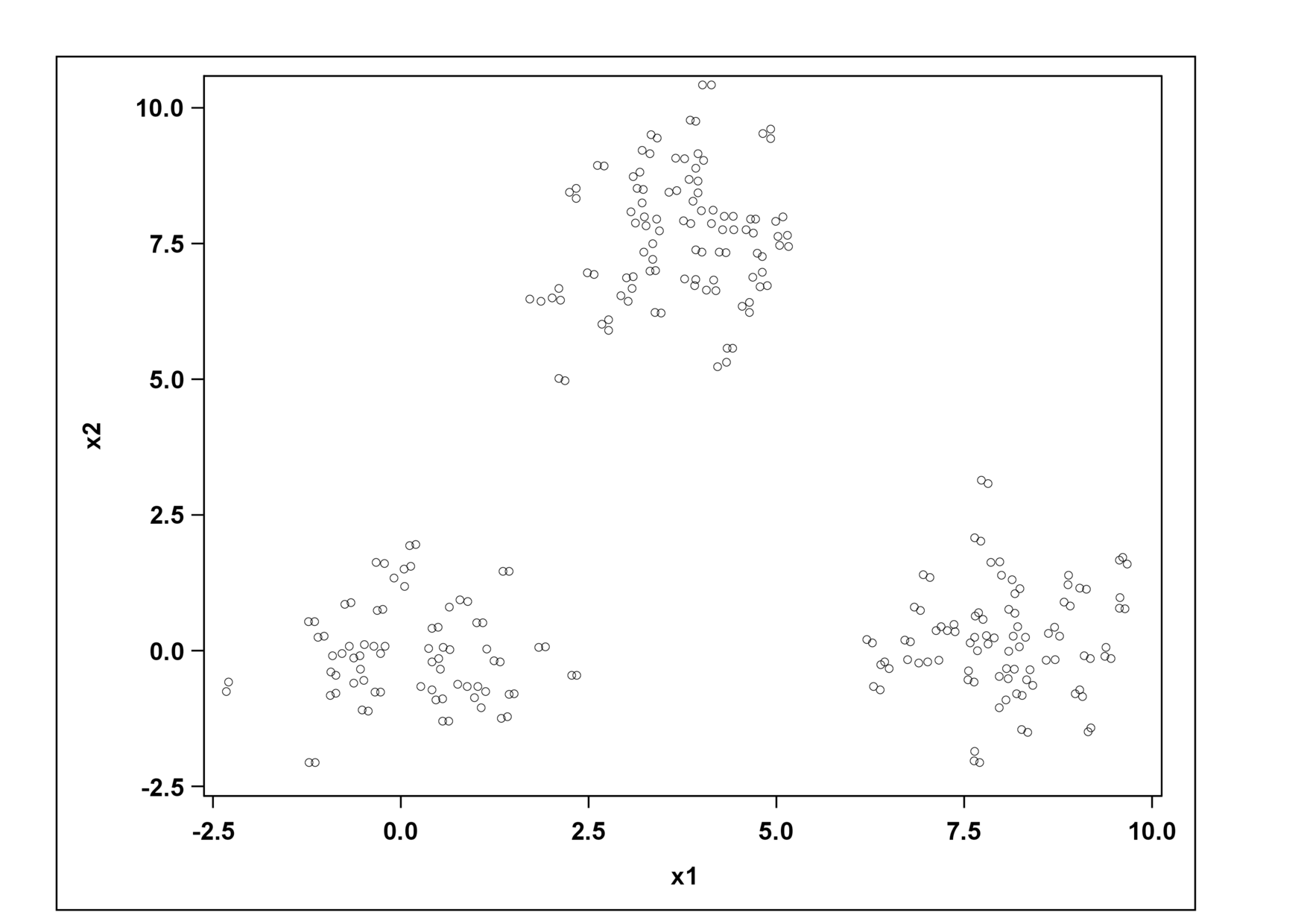}}
\subfigure[s=0.4]{\label{fig:image9}\includegraphics[width=2.5in]{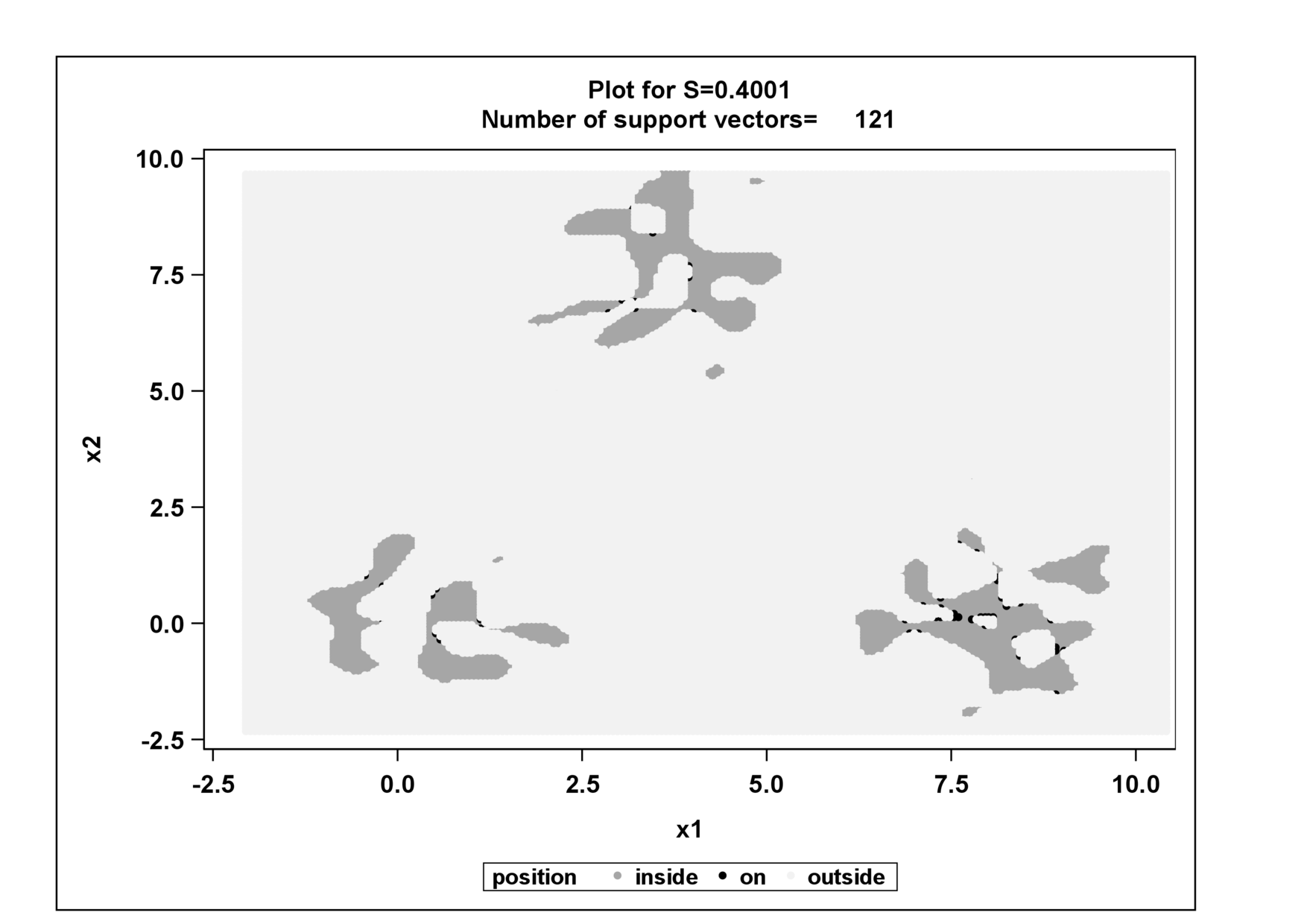}}
\subfigure[s=1.1]{\label{fig:image10}\includegraphics[width=2.5in]{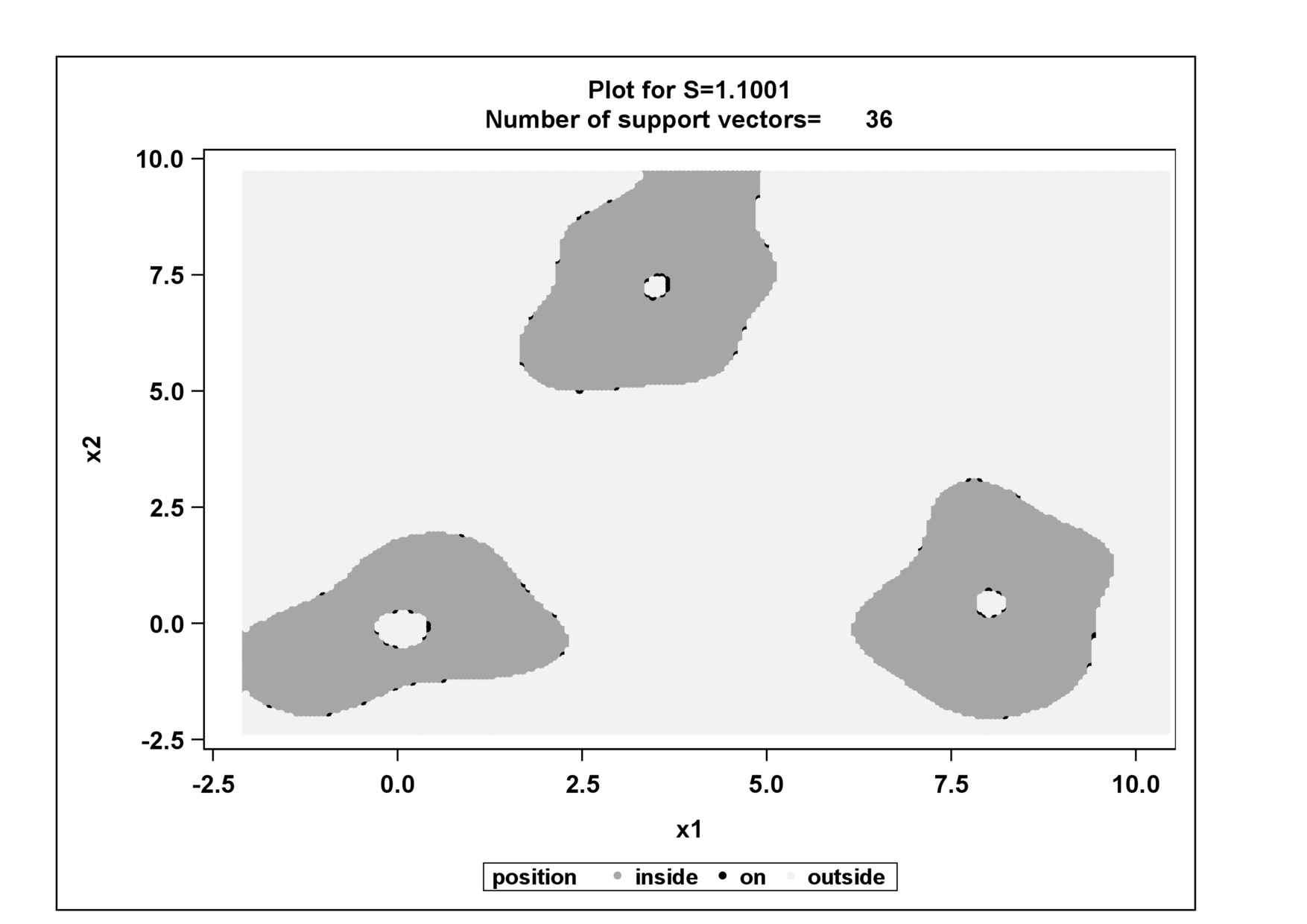}}
\subfigure[s=3.5]{\label{fig:image11}\includegraphics[width=2.5in]{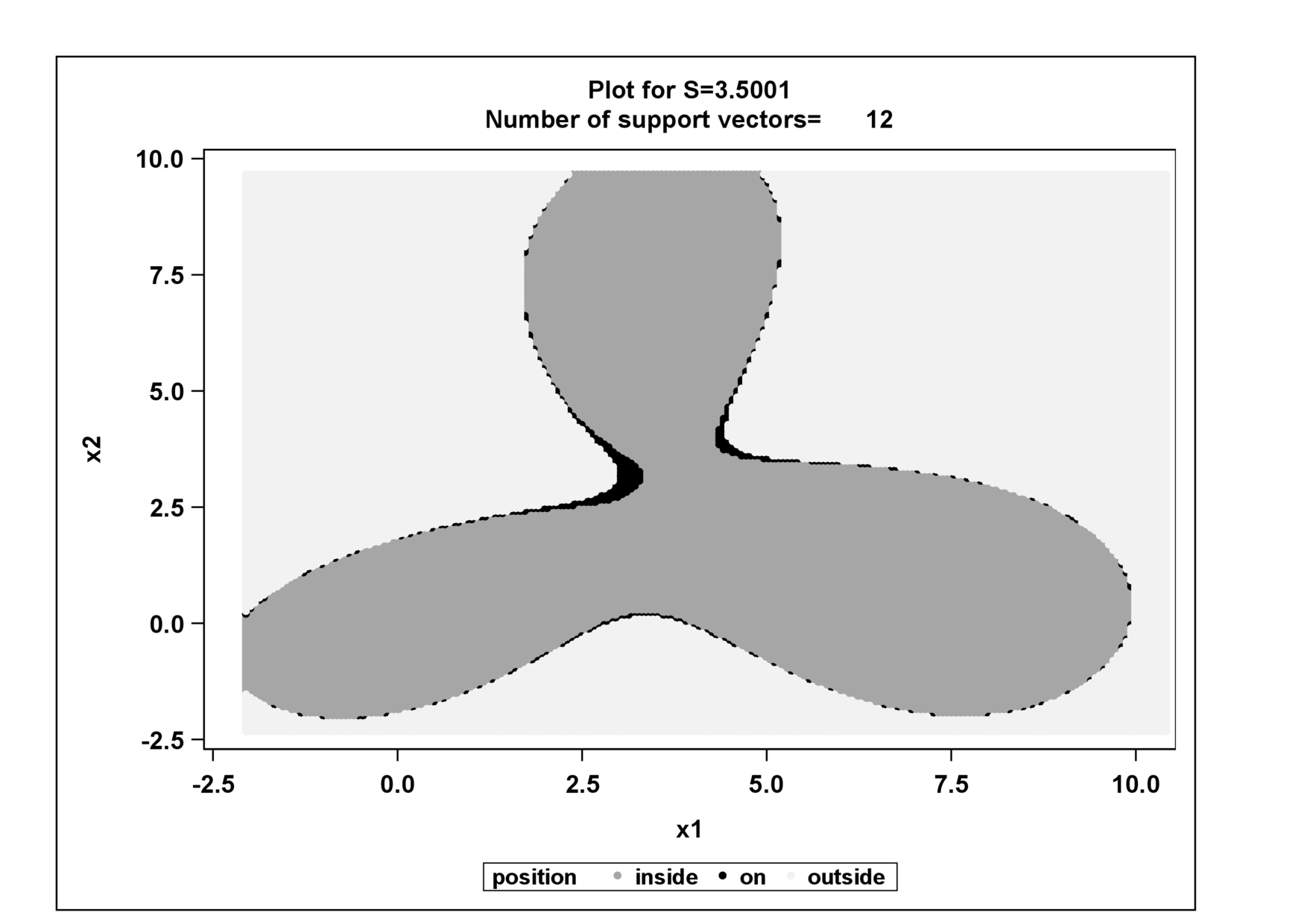}}
\caption{Data boundary for three-cluster data.Fig (b) thru (d) show results of scoring on a 200x200 data grid. Light gray color indicates outside points, dark gray color indicates inside points and black color indicates support vectors. }\label{fig:image_9}
\end{figure}

We performed similar experimentation using star-shaped data and a data set with three distinct data clouds. The three cluster data was obtained from the \textit{SAS/STAT User's guide}  \cite{institute2015sas}.
Figure \ref{fig:image_9}(a) shows a scatter plot of this latter data set. 

Similar to the banana-shaped data, we trained the three-cluster data set varying $s$ from 0.001 to 8 in increments of 0.05. Scoring was performed on a 200x200 data point grid to confirm the shape of the data boundary. Scoring results for select values of $s$ are provided in figure \ref{fig:image_9}(b-d).

\begin{figure}
    \centering
    \includegraphics[width=0.5\textwidth]{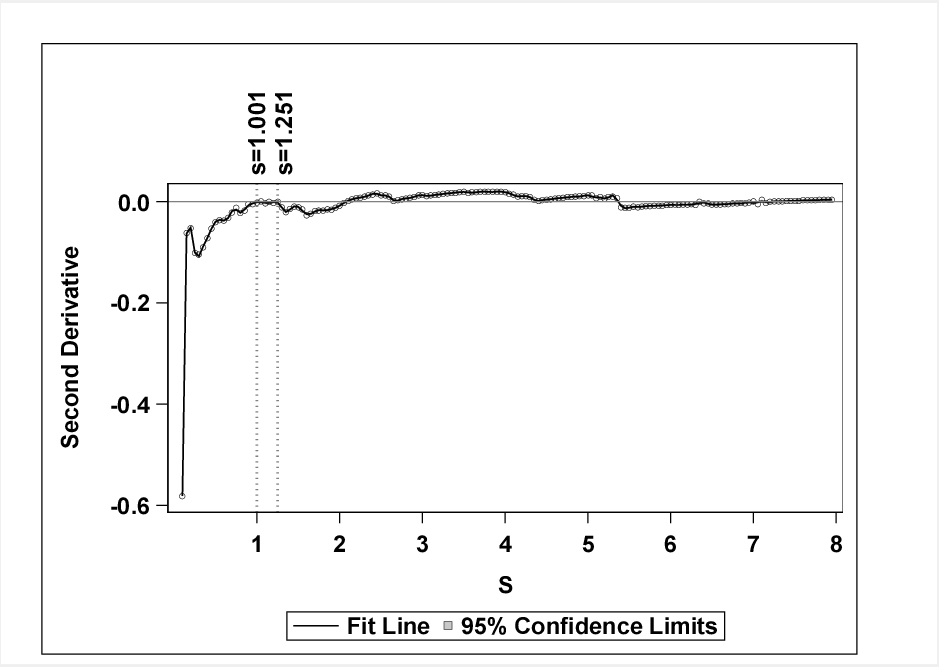}
    \caption{Penalized B-spline fit for second derivative: Three-cluster data}
    \label{fig:image_12}
\end{figure}

\begin{figure}
\subfigure[{\normalsize Scatterplot of star-shaped data}]{\label{fig:image13}\includegraphics[width=2.5in]{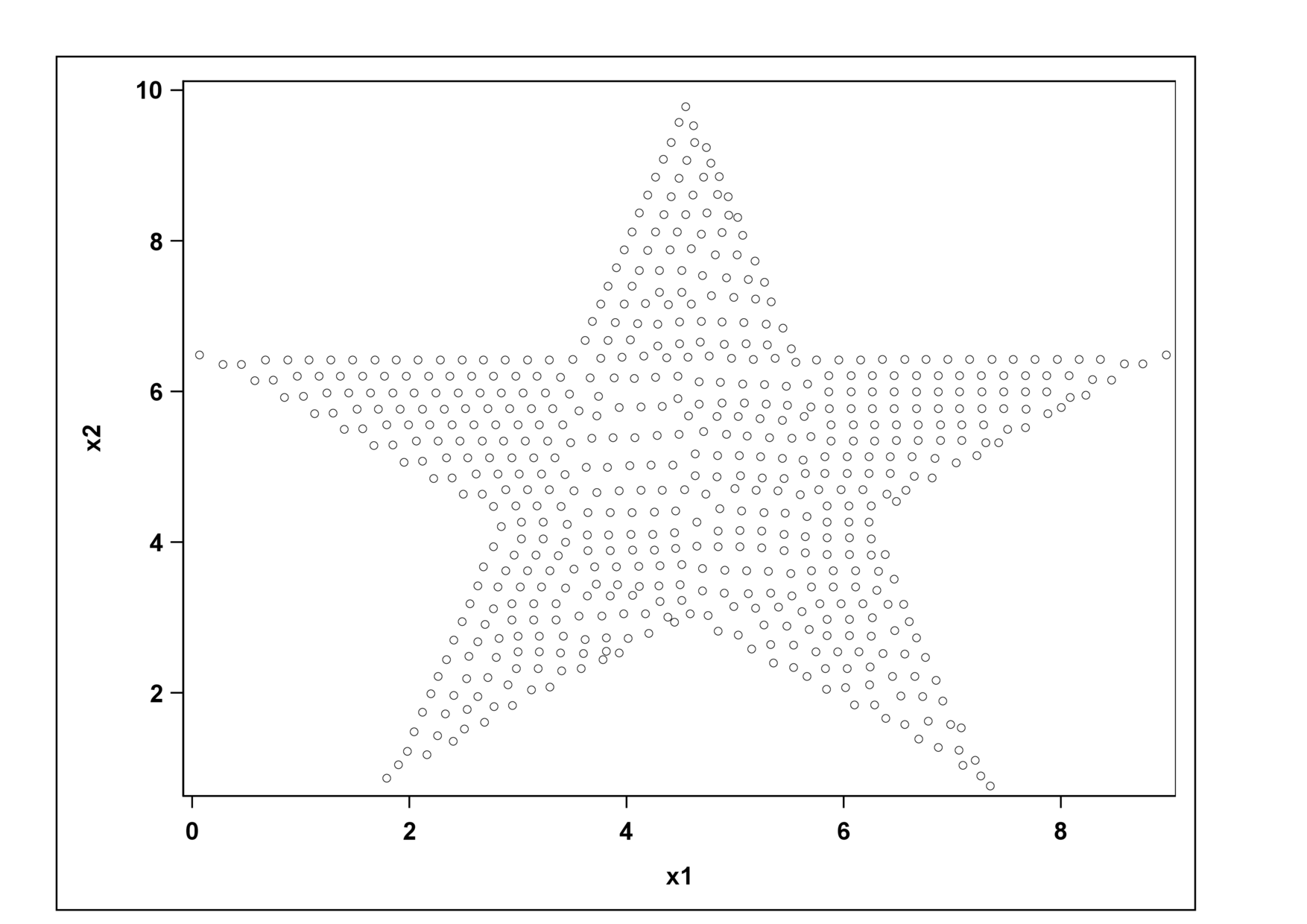}}
\subfigure[{\large s=0.2}]{\label{fig:image14}\includegraphics[width=2.5in]{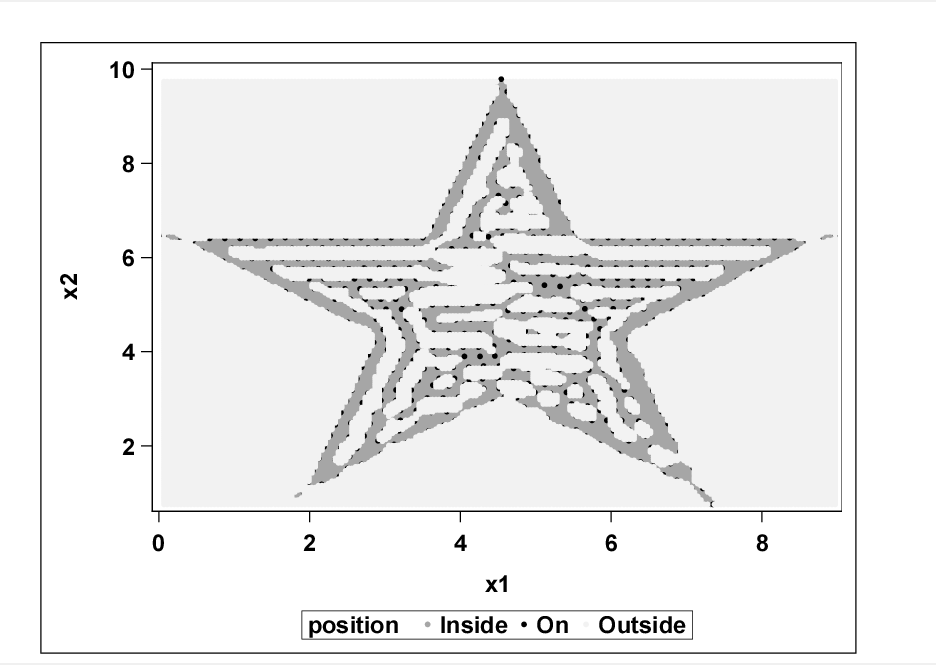}}
\subfigure[{\large s=0.9}]{\label{fig:image15}\includegraphics[width=2.5in]{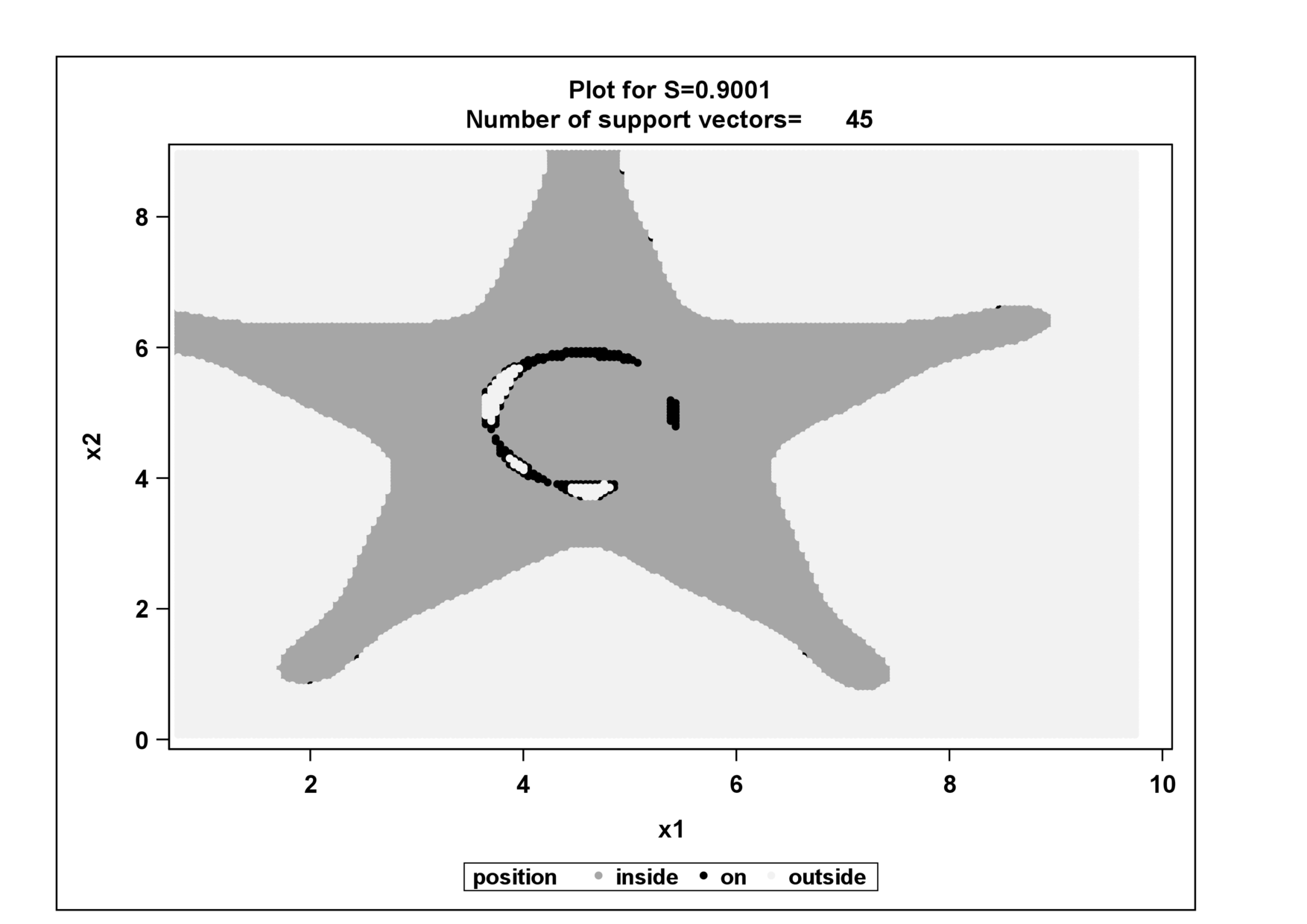}}
\subfigure[{\large s=2.3}]{\label{fig:image16}\includegraphics[width=2.5in]{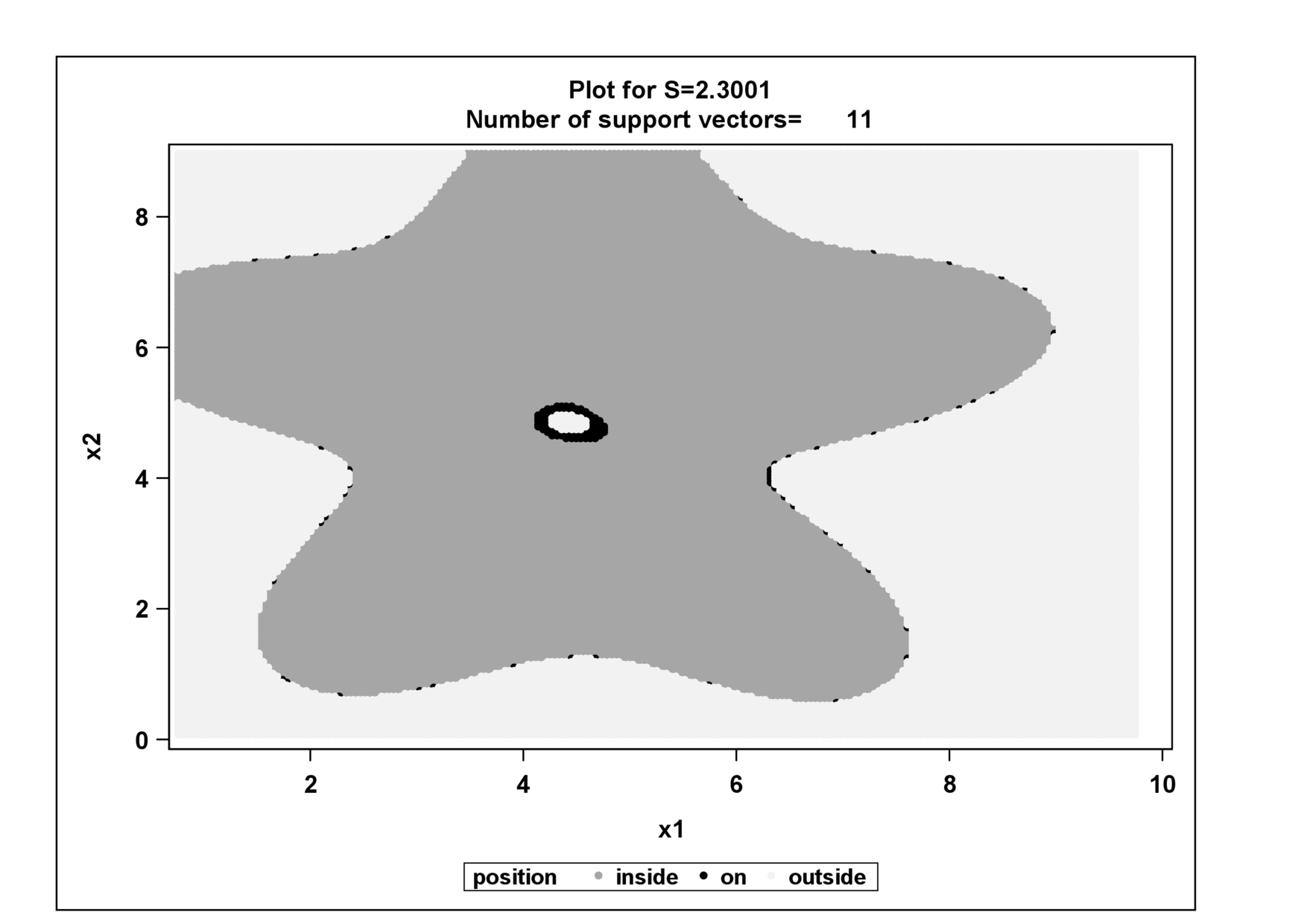}}
\caption{Data boundary for star-shaped data. Fig (b) thru (d) show results of scoring on a 200x200 data grid. Light gray color indicates outside points, dark gray color indicates inside points and black color indicates support vectors.}\label{fig:image_14}
\end{figure}

Figure \ref{fig:image_12} shows the second derivative of  $ V^{*}(s) $ with respect to $s$ for the three-cluster data. The results are similar to the banana-shaped data.  For $ s$ in $ [1.0,1,25] $, the second derivative is close to zero indicating this is the first set of critical points. For these values, high quality data boundaries were obtained. To illustrate, the data boundary using $s$=1.1 is shown in Figure \ref{fig:image_9}(c). The boundary captures the essential geometric properties of the three-cluster data especially in comparison to any other values of $s$ outside the first critical value interval (see Figure \ref{fig:image_12}(b) and (d)). 
Next, we conducted our experiments with a star-shaped data set. Figure \ref{fig:image_14}(a) shows the scatter plot of these data. This data set was trained  using values of $s$ from 0.001 to 8 in increments of 0.05. Scoring was performed on a 200x200 data point grid to confirm the shape of data boundary. Scoring results for select values of $s$ are provided in Figure \ref{fig:image_14}.

\begin{figure}
    \centering
    \includegraphics[width=0.5\textwidth]{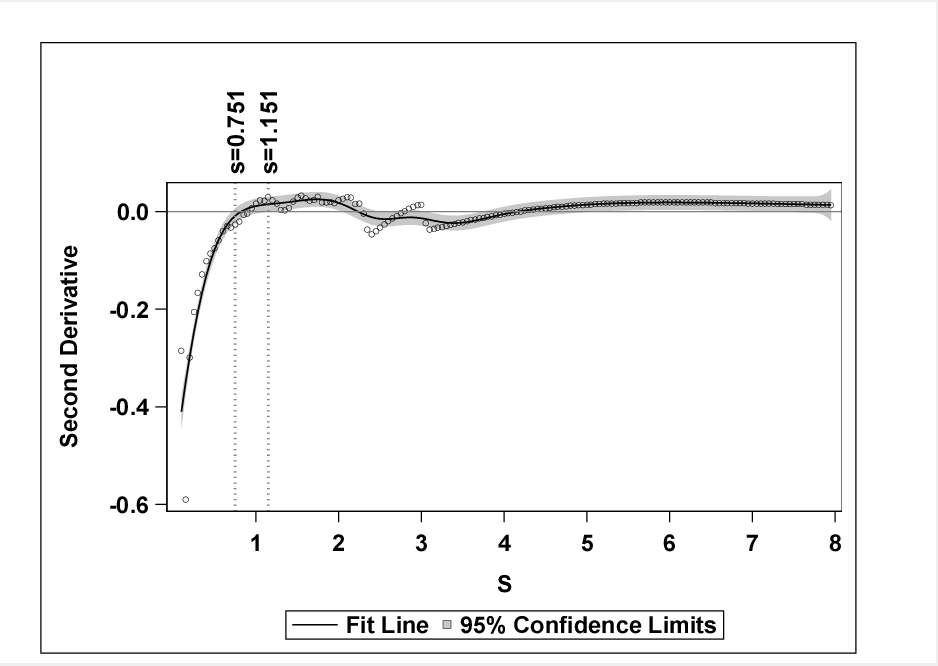}
    \caption{Penalized B-spline fit for second derivative: star-shaped data}
    \label{fig:image_17}
\end{figure}

Figure \ref{fig:image_17} shows the second derivative of the optimal value of the objective function ($ V^{*}(s)$) with respect to $s$ for the star-shaped data.  Between $s$=0.75 and $s$=1.15, for the first time, the second derivative is close to zero for the first time; the first, first derivative critical point is reached. A data boundary of good quality is observed at values of s between 0.75 and 1.15 (see Figure \ref{fig:image_14}(c)); the data boundary captures the essential geometric properties of the data especially when compared to any other values of $s$ (for examples, see Figure \ref{fig:image_14}(b) and (d)).

We tried our analysis on data sets with diverse geometrical shapes. For all data sets, the fact that a good quality data boundary can be obtained using value of $s$ from the first set of critical points of the first derivative of $ V^{*}(s) $, provides the empirical basis for our method. 

\section{Analysis of High Dimensional Data}
\label{pd}
Section \ref{crit} illustrated the value of using the Peak criterion to select $s$ for different two-dimensional data sets. For such data sets a good value of $s$ could be visually judged. Next, we want to test the criterion on higher dimensional data sets, where visual feedback about a good value of $s$ is not possible. Instead, we see how the Peak criterion $s$ values fare based on a measure used to assess model quality when labeled data are available. This criterion, known as the $F_{1}$-measure \cite{zhuang2006parameter} is defined as follows:
\begin{equation}  
	F_{1}=\dfrac{2\times \text{Precision}\times \text{Recall}}{\text {Precision}+\text {Recall}},
\end{equation}
where:
\begin{align}
	\text {Precision}=\dfrac{\text{true positives}}{\text{true positives} + \text{false positives}}\\
	\text {Recall}=\dfrac{\text{true positives}}{\text{true positives} + \text{false negatives}}.
\end{align} 
We chose the $F_{1}$-measure because it is a composite measure that takes into account both Precision and Recall. Models with higher values of the $F_{1}$-measure are assumed to provide a better fit.  \\

\subsection{Analysis of Shuttle Data}
The first higher dimensional data set we analyze is the Statlog (shuttle) data \cite{Lichman:2013}. It consists of nine numeric attributes and one class attribute. Out of 58,000 total observations, 80\% of the observations belong to class one. A random sample of 2000 observations belonging to class one, was selected for training. Scoring was performed to determine if the model could accurately classify an observation as belonging to class one. The SVDD model was trained and subsequently scored for values of $s$ ranging from 1 to 100 in increments of 1. For each value of $s$ the model performance was quantified using the $F_{1}$-measure.  

\begin{figure}
 	\centering
 	\includegraphics[width=0.5\textwidth]{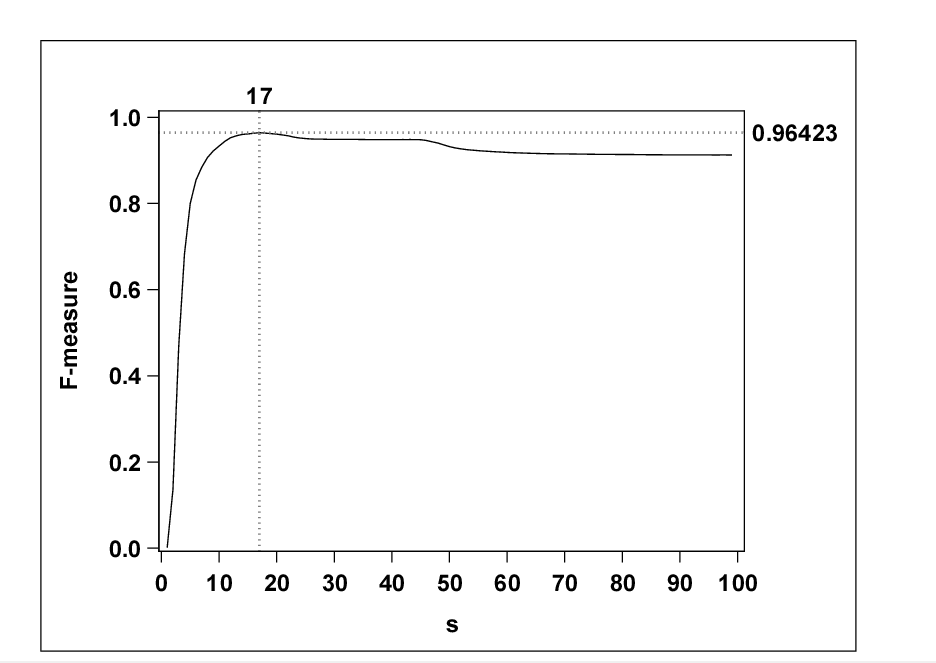}
 	\caption{Bandwidth parameter vs. $F_{1}$ measure: shuttle data}
 	\label{fig:image_30}
 \end{figure}

\begin{figure}
  	\centering
  	\includegraphics[width=0.5\textwidth]{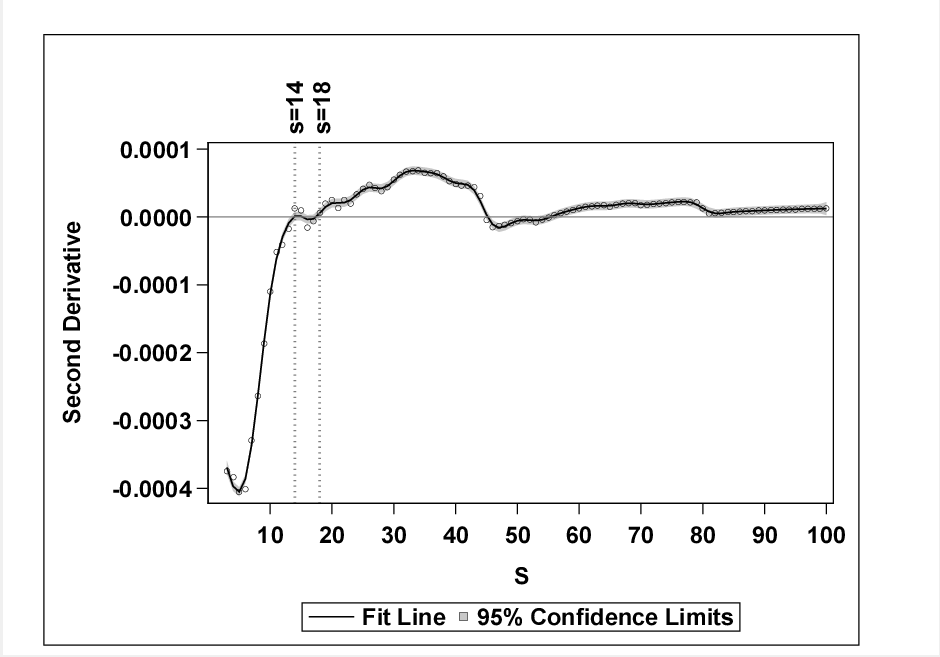}
  	\caption{Penalized B-spline fit for second derivative: shuttle data}
  	\label{fig:image_31}
  \end{figure}

The plot of the $F_{1}$-measure versus $ s $ is shown in Figure \ref{fig:image_30}. A maximum value of $F_{1}$-measure is obtained at $s$=17. Interestingly, the function is quite flat around $s$=17. In fact, the $F_{1}$-measure is very similar for $ s $ in [15,20]. Figure \ref{fig:image_31} shows the plot of the second derivative of optimal value of objective function with respect to $s$ plotted against $s$ for this data. The values of $s$ between 14 and 18, where the second derivative is nearly zero represents the first set of critical points. The fact that value of $s$=17 obtained using the $F_{1}$-measure belongs to the set [14,18], obtained by the Peak criterion, provides the empirical evidence that Peak criterion works successfully with higher dimensional data.

\subsection{Analysis of Tennessee Eastman (TE) Data}
In this section we provide results of our experiments with the higher dimensional Tennessee Eastman  data.  The data were generated using MATLAB simulation code \cite{Ricker:2002} which provides a model of an industrial chemical process\cite{downs1993plant}. The data were generated for normal operations of the process and twenty faulty processes. Each observation consists of 41 variables out of which 22 are measured continuously, on an average of every 6 seconds, and the remaining 19 are sampled at a specified interval either every 0.1 or 0.25 hours. We created our analysis data set using the simulated normal operations data for the first 90 minutes, followed by data corresponding to faults 1 through 20. A random sample of 200 observations belonging to normal operations, was selected for training. Scoring was performed on the remaining observations to determine if the model could accurately classify an observation as belonging to normal operations of the process. The SVDD model was trained and subsequently scored for values of $s$ ranging from 1 to 100 in increments of 1. For each value of $s$ the model performance was quantified using the $F_{1}$-measure.  

\begin{figure}
 	\centering
 	\includegraphics[width=0.5\textwidth]{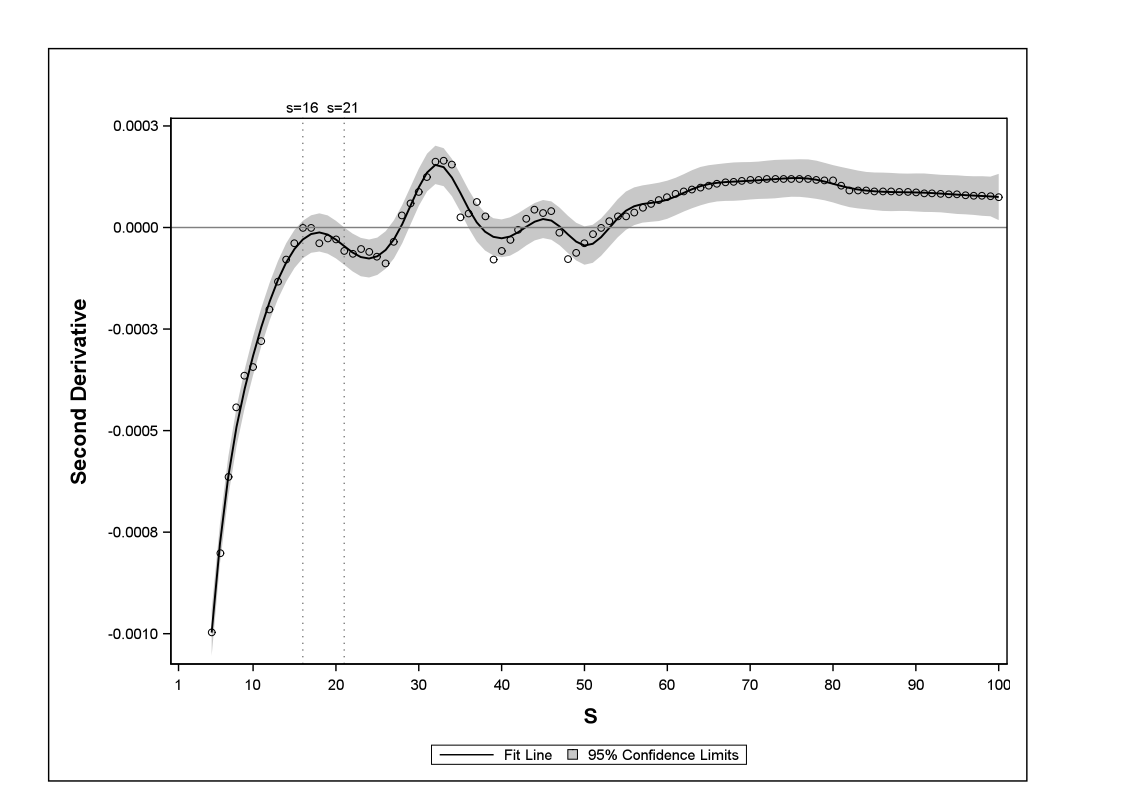}
 	\caption{Penalized B-spline fit for second derivative: Tennessee Eastman data}
 	\label{fig:image_40}
 \end{figure}

\begin{figure}
 	\centering
 	\includegraphics[width=0.5\textwidth]{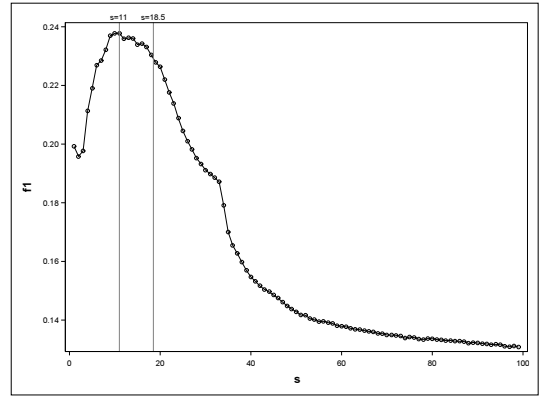}
 	\caption{Bandwidth parameter vs. $F_{1}$ measure: Tennessee Eastman data}
 	\label{fig:image_41}
 \end{figure}

Figure \ref{fig:image_40} shows the plot of the second derivative of $ V^{*}(s)$ with respect to $s$ plotted versus $s$. The values of $s$ between 16 and 21, where the second derivative is nearly zero, represent the first set of critical points. The plot of the $F_{1}$-measure versus $ s $ is shown in  Figure \ref{fig:image_41}. A maximum value of $F_{1}$-measure (0.2378) is obtained at $s$=11. The value of $F_{1}$-measure at the midpoint of the $s$ range suggested by the Peak criteria is 0.2291. The fact that the $F_{1}$-measure value for the $s$ value suggested by the Peak criteria is about 95\% of the maximum value of the  $F_{1}$-measure, provides more empirical evidence that Peak criterion works successfully with higher dimensional data.

\section{Simulation Study}
\label{ss}
In this section we measure the performance of Peak criterion when it is applied to randomly generated polygons. Given
the number of vertices, $k$,we generate the vertices of a randomly generated polygon in the anticlockwise sense as $r_1
\exp i \theta_{(1)}, \dots, r_k \exp i \theta_{(k)}.$ Here $\theta_{(1)} = 0$ and $\theta_{(i)}$'s for $i = 2,\dots,n$ are the order statistics of an i.i.d sample
uniformly drawn from $(0,2\pi).$ The $r_i$'s are uniformly drawn from an interval $[\text{r}_{\text{min}},\text{r}_{\text{max}}].$ 

\begin{figure}
\begin{tabular}{cc}

	\includegraphics[width=2.5in]{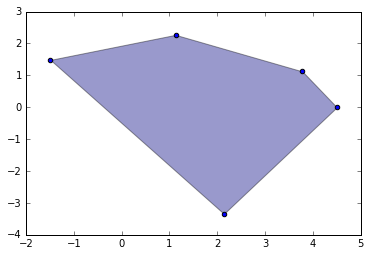} \\
     (a) Number of Vertices = 5 \\
    \includegraphics[width=2.5in]{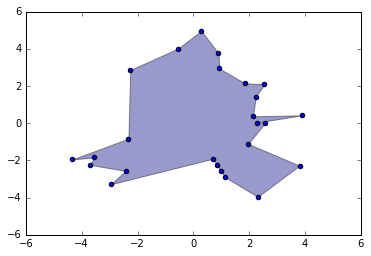} \\
	 (b) Number of Vertices = 25\\

\end{tabular}
\caption{Random Polygons}\label{fig:image_44}
\end{figure}

For this simulation we chose $\text{r}_{\text{min}}=3$ and
$\text{r}_{\text{max}}=5$ and varied the number of vertices from $5$ to $30$.We generated $20$ random polygons for each
vertex size. Having determined a polygon we randomly sampled $600$ points uniformly from the interior of the
polygon and used this sample to determine a bandwidth using the Peak criterion. Figure \ref{fig:image_44} shows two random polygons.

However since we can easily determine if a point lies in the interior of a polygon
we can also use cross-validation to determine a good bandwidth value.  
To do so, we found the bounding rectangle of each of the polygons and divided it into a $200 \times 200$ grid. We then
labeled each point on this grid as an ``inside'' or an ``outside'' point. We then fit SVDD on the sampled data and scored the points on this grid for different
values of $s$ and choose that value that value of $s$ that maximized the $F_1$-measure.

\begin{figure}
	\centering
	\includegraphics[width=0.55\textwidth]{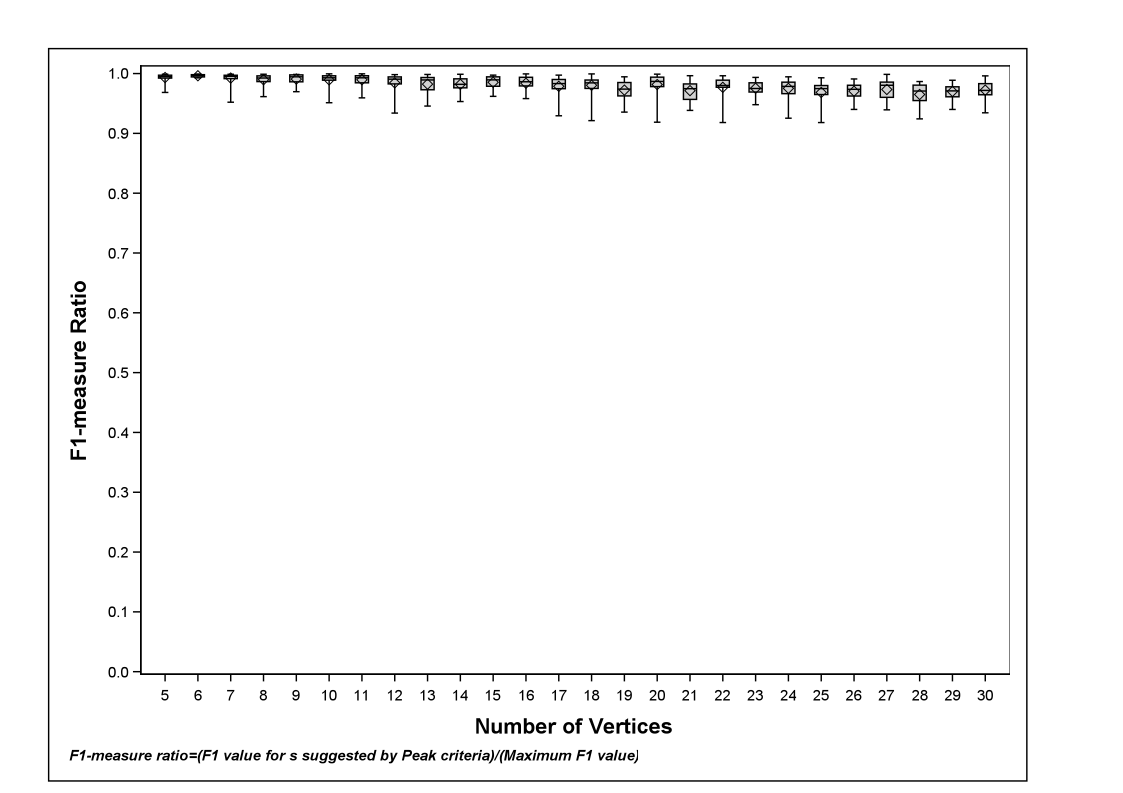}
	\caption{Box-whisker plot: Number of vertices vs. $F_1$ measure ratio }
	\label{fig:image_42}
\end{figure}

The performance of the Peak criterion can measured by the $F_1$-measure ratio defined as $F_{\text{peak}}/F_{\text{best}}$ where
$F_{\text{peak}}$ is the $F_1$-measure obtained when the value suggested by the Peak method is used, and
$F_{\text{best}}$ is the best possible value of $F_1$-measure over all values of $s$. A value close to 1 wll indicate
that Peak criterion is competitive with cross-validation. We have $20$ values of this ratio for each vertex size.\\
The Box-whisker plot in Figure \ref{fig:image_42} summarizes the simulation study results. The x- axis shows the number
of vertices of the ploygon and y-axis shows the $F_1$-measure ratio. The bottom and the top of the box shows the first and the third quartile values. The ends of the whiskers represent the minimum and the maximum value of the $F_1$-measure ratio. The diamond shape indicates the mean value and the horizontal line in the box indicates the second quartile. The plot shows that $F_1$-measure ratio is greater than 0.9 across all values of number of vertices. The $F_1$ measure ratio in the top three quartiles is greater than 0.95 across all values of the number of vertices.As the complexity of the ploygon increases with increase in number of vertices, we observed that the spread of $F_1$-measure ratio also increased. The fact that $F_1$-measure ratio is always close to 1, provides necessary evidence that the Peak criterion generalizes across different training data sets.  

\section{Related Work} 
\label{rw}

\begin{figure}[ht!]
	\begin{tabular}{cc}
	   \includegraphics[width=2.5in]{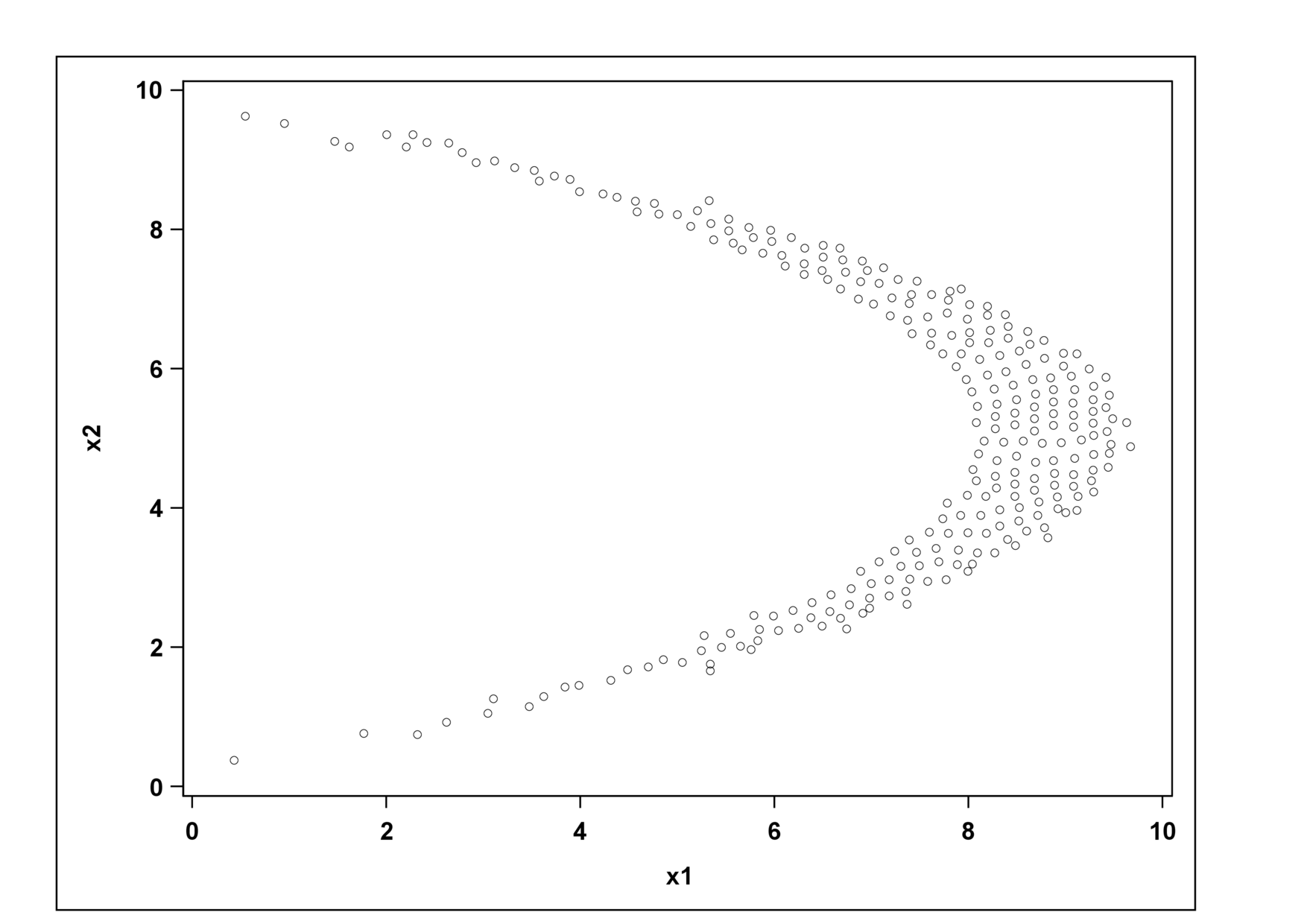}\\
	    (a) Original data\\
		\includegraphics[width=2.5in]{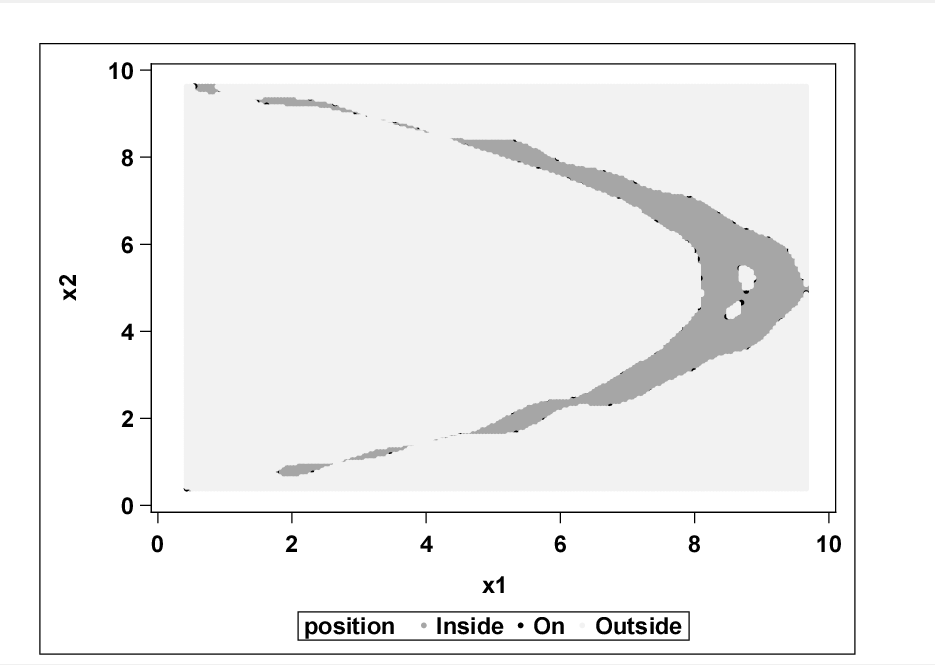}\\
        (b) CV\\
        \includegraphics[width=2.5in]{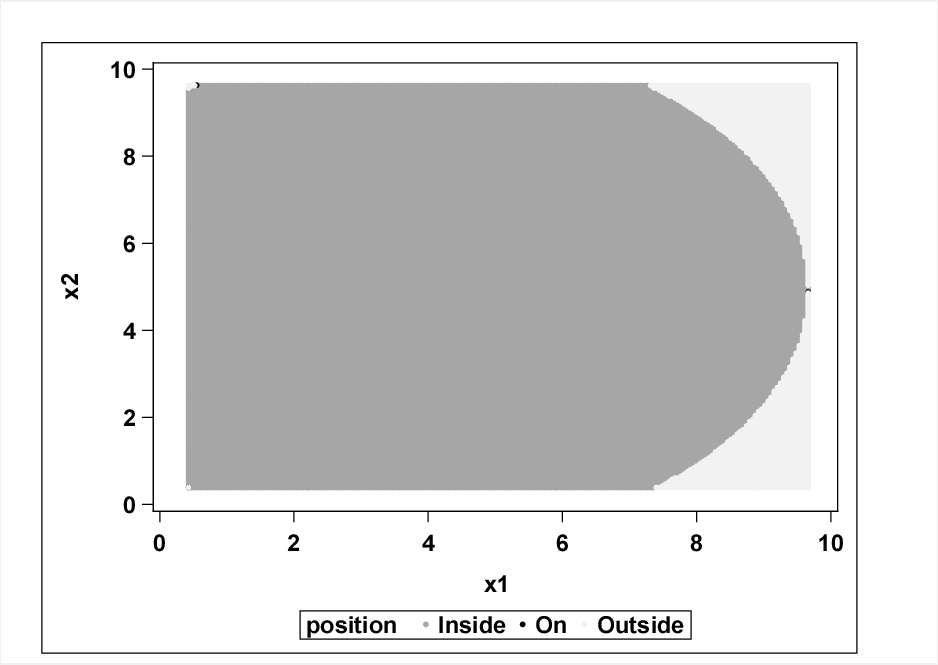}\\
		(c) MD\\
		\includegraphics[width=2.5in]{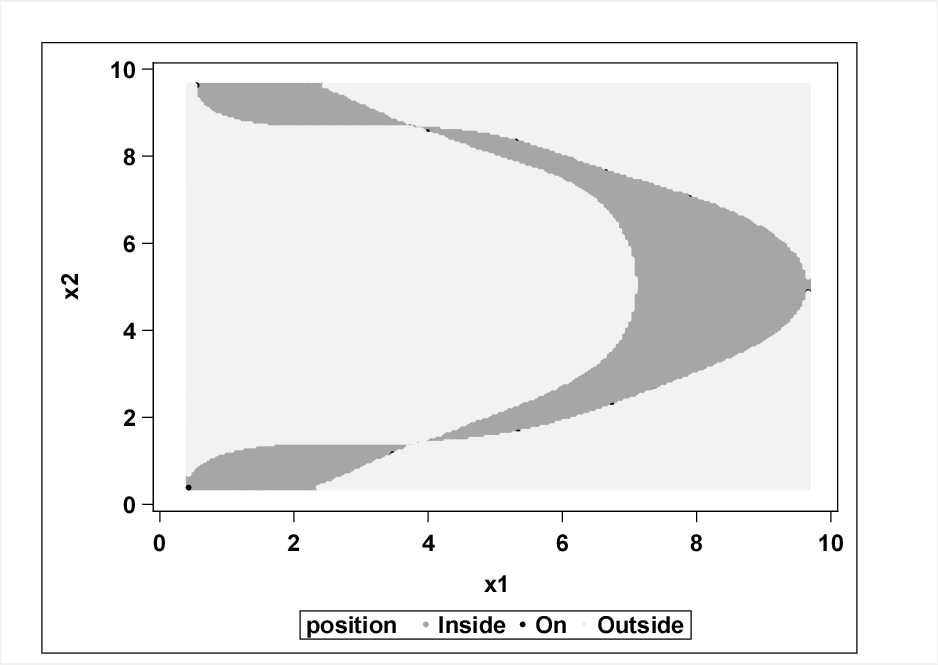}\\
        (d) DFN\\
        \includegraphics[width=2.5in]{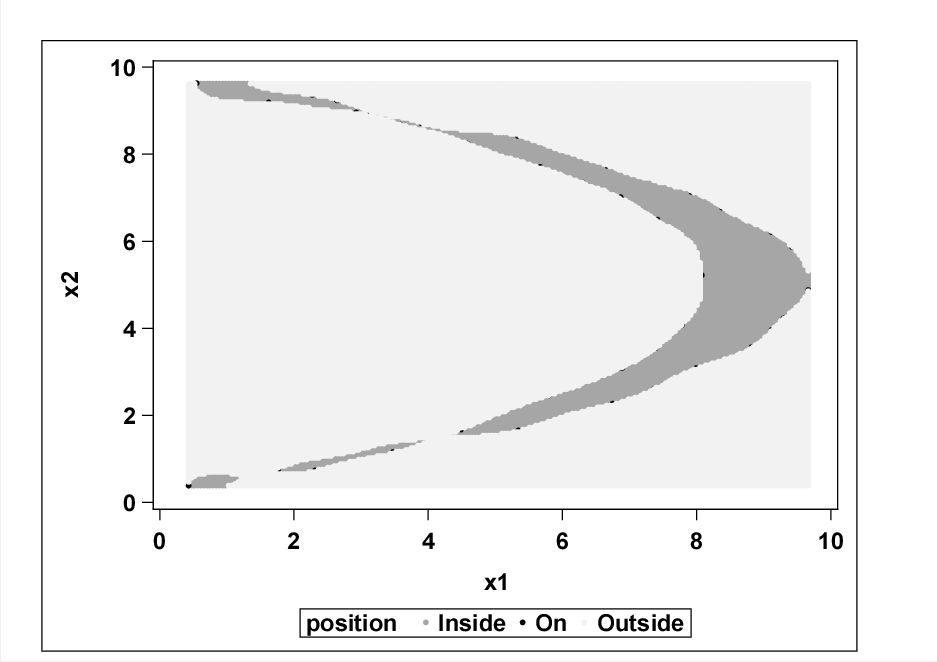}\\
		(e) Peak\\
	\end{tabular}
	\caption{Banana-shaped data}\label{fig:image_18}
\end{figure}
\begin{figure}[ht!]
	\begin{tabular}{cc}
	    \includegraphics[width=2.5in]{image8.png}\\
	    (a) Original data\\
		\includegraphics[width=2.5in]{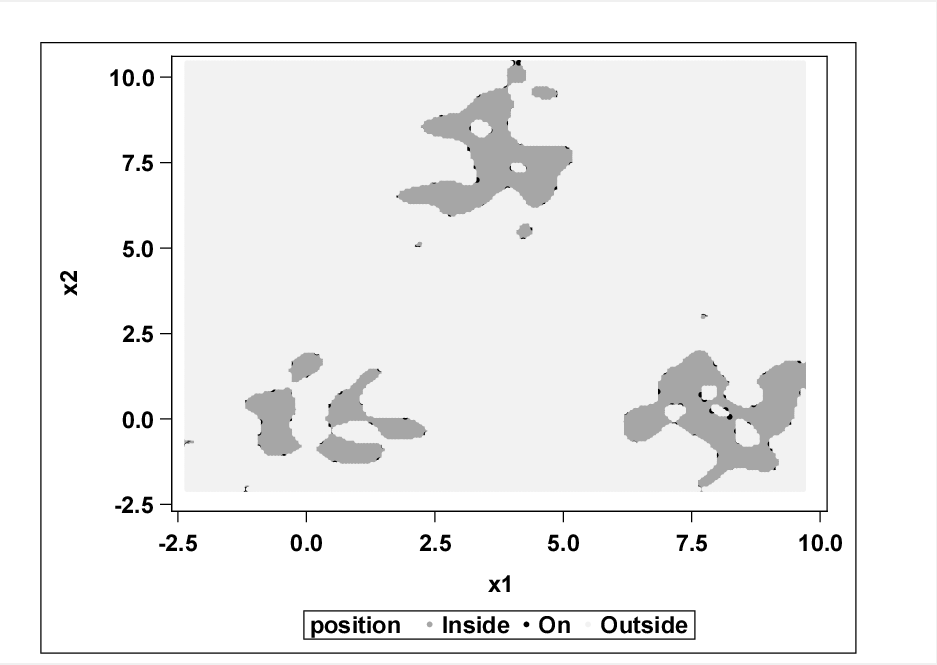} \\
                               (b) CV\\
        \includegraphics[width=2.5in]{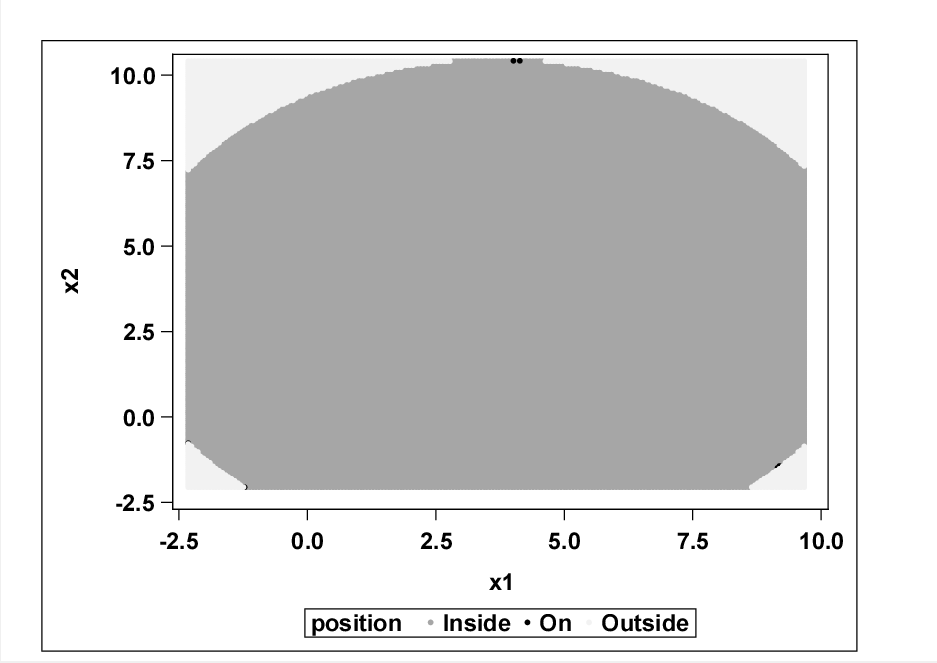} \\
		(c) MD\\
		\includegraphics[width=2.5in]{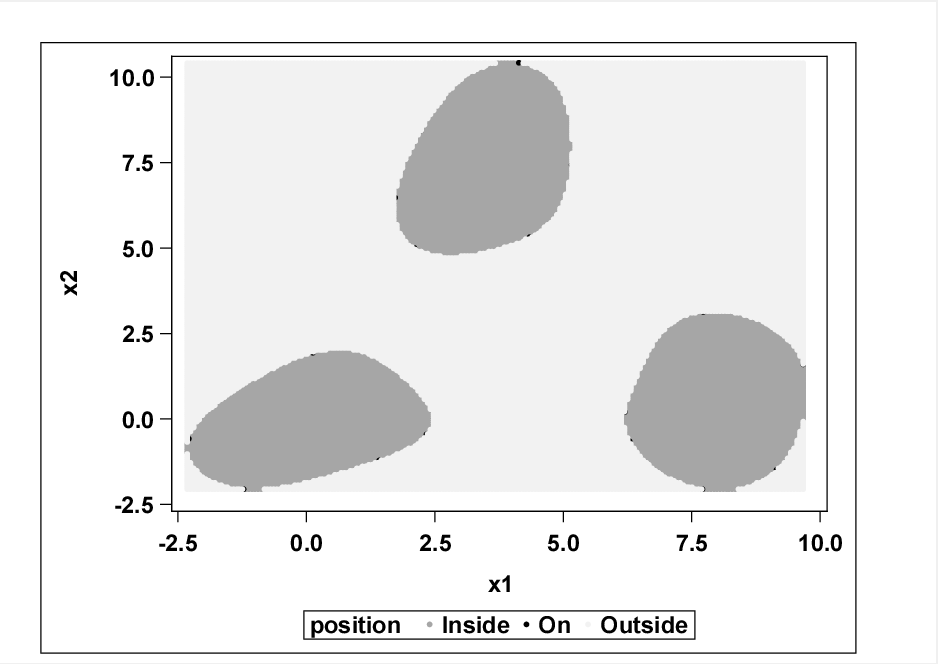}\\ 
        (d) DFN\\
       \includegraphics[width=2.5in]{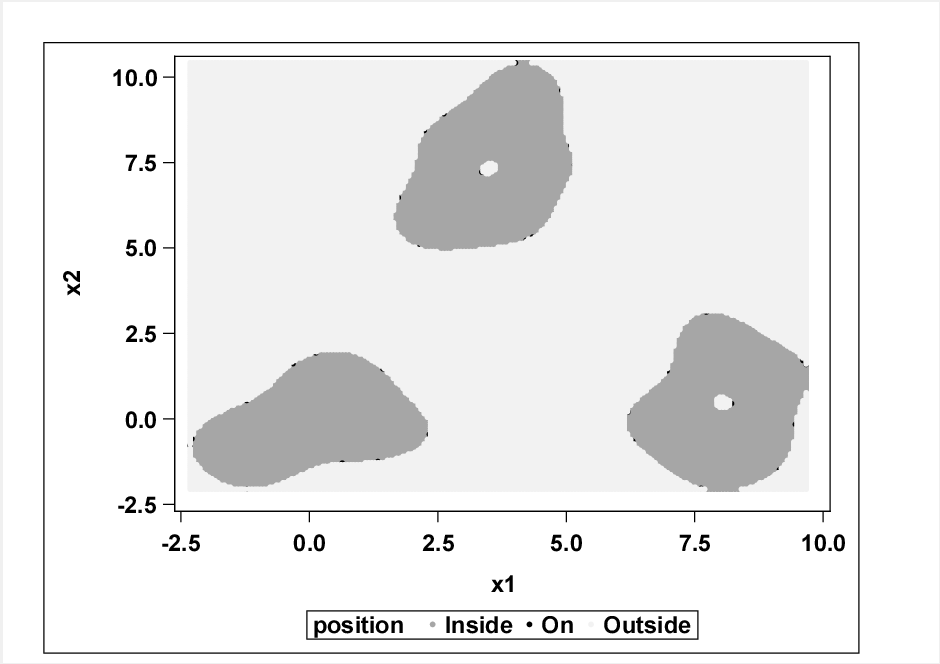} \\
	    (e) Peak
	\end{tabular}
	\caption{Three-cluster data}\label{fig:image_19}
\end{figure}
\begin{figure}[ht!]
	\begin{tabular}{cc}
	   \includegraphics[width=2.5in]{image13.png}\\
	   (a) Original data\\
		\includegraphics[width=2.5in]{image14.png}\\
       (b) CV\\
        \includegraphics[width=2.5in]{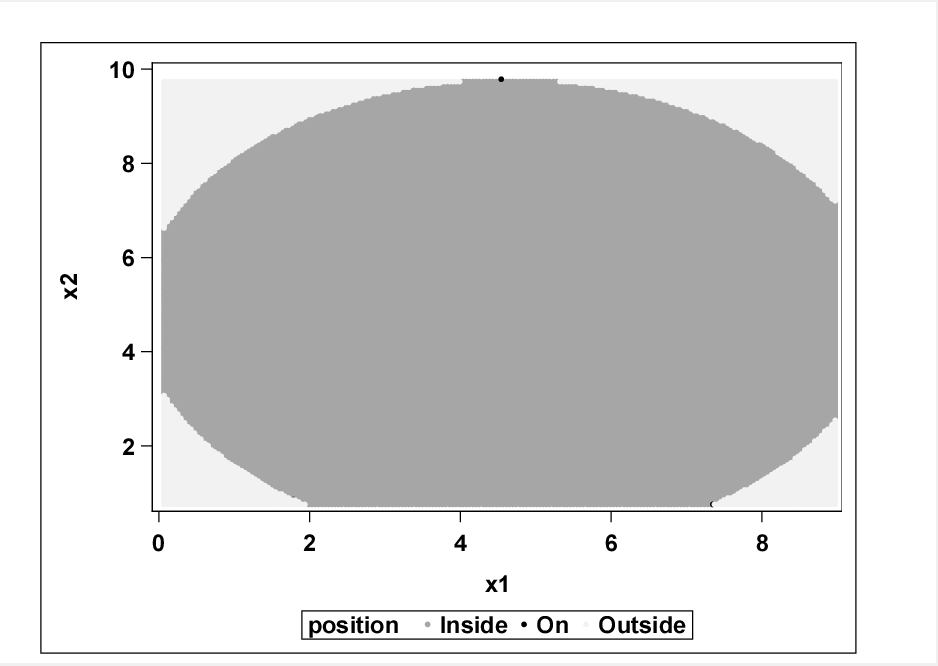}\\
		(c) MD\\
		\includegraphics[width=2.5in]{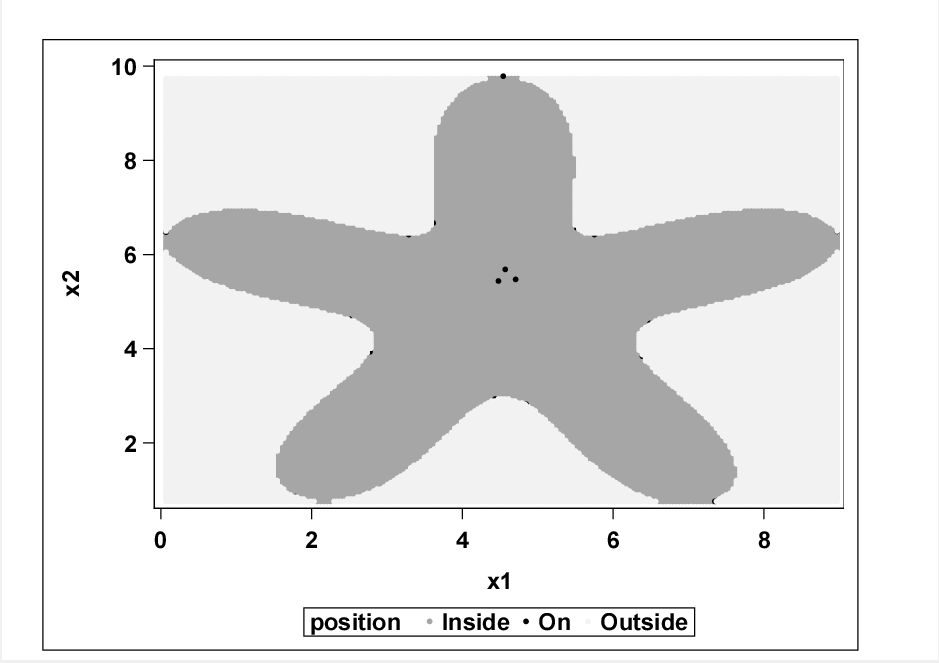}\\ 
        (d) DFN\\
         \includegraphics[width=2.5in]{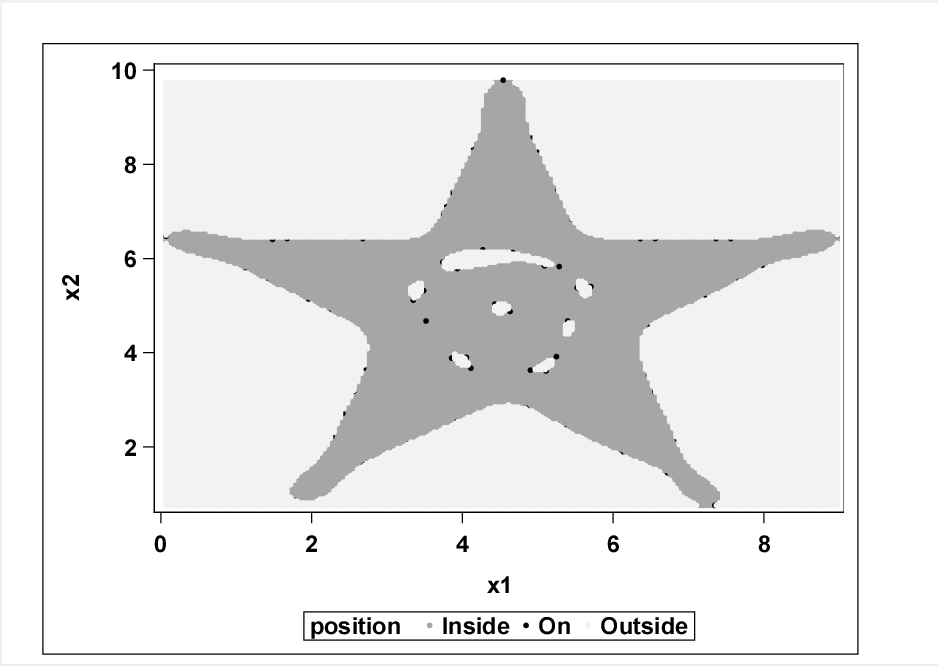} \\
		(e) Peak \\
	\end{tabular}
	\caption{Star-shaped data}\label{fig:image_20}
\end{figure}

 In support vector machines, cross-validation is a widely used technique for selecting the Gaussian bandwidth parameter \cite{hastie2009unsupervised}. Cross-validation requires training data that belongs to multiple classes. 
 Hence, unless a good sample for normal class and outlier class is available, cross-validation is not a feasible technique for selecting Gaussian bandwidth parameter value in SVDD.\\
 The Peak criterion is an unsupervised method that works on single class data. In this section, performance of the Peak criterion is compared against unsupervised methods for selecting Gaussian bandwidth parameter value published in the literature. \\
 \\
\textbf{Method of Coefficient of Variation (CV)} \cite{evangelista2007some}:\\
Selects a value of $s$ that maximizes the coefficient of variation of the kernel matrix. 
\begin{equation}  
  CV=\dfrac{\text{Var}}{\text{Mean}+\epsilon}
\end{equation}
where:\\
Var and Mean are variance and mean of the non-diagonal entries of the kernel matrix,\\
$\epsilon$ is a small value to protect against division by zero or round-off error. In our CV method computations, we set the value of $\epsilon$ to 0.000001.  \\
\\
\textbf {Method of Maximum Distance (MD)} \cite{khazai2011anomaly}: \\
Obtains a value of $s$ based on maximum distance between any pair of points in the training data.
\begin{equation}
 s=\frac{d_{max}}{\sqrt{-ln(\delta )}} 
\end{equation}
where:\\
$d_{max}=max{ \|x_i - x_j\|^2 } $: maximum distance between any two pairs of points,\\
$\delta =\frac{1}{n(1-f)+1}$,\\
$n:$Number of observation in training data,\\
$f:$ the expected outlier fraction. In our MD method computations, we set the value of $f$ to 0.001 \\
\\
\textbf{Method of Distance to the Farthest Neighbor (DFN)} \cite{xiao2014two}:\\
Uses distances of the training data points to their farthest neighbors and distances to their nearest neighbors. The optimal value of $s$ is obtained by maximizing the following objective function:\\
\begin{equation}
f_{0}(s)= \frac{2}{n}\sum_{i=1}^{n}max_{j\neq i}k(x_{i},x_{j})-\frac{2}{n}\sum_{i=1}^{n}min_{j}k(x_{i},x_{j}).\\
\end{equation}
where:\\
$n:$ number of observations in training data,\\
$k(x_{i},x_{j}):$ kernel distance between observations $i$ and $j$.\\

We calculated the values of s for the banana-shaped, three-cluster, and star-shaped data using the CV, MD and DFN method. Table~\ref{table:t1} summarizes these results and also provides the value of $s$ obtained using the Peak criteria.

The scoring results using values of $s$ recommended by above methods are illustrated in Figure \ref{fig:image_18},  Figure \ref{fig:image_19} and Figure \ref{fig:image_20}. For all three data sets, when compared against existing methods, the Peak criterion clearly provides a data boundary of best quality. The method of Coefficient of Variation also provides a data boundary of fairly good quality. 

\begin{table}[h!]
\centering
 \begin{tabular}{||c c c c c||} 
 \hline
 Data & CV & MD & DFN & Peak \\
 \hline\hline
 Banana  & 0.5 & 46 & 1.99 & 0.4 to 1.1 \\ 
 Three-cluster & 0.55 & 77 & 1.98 & 1.0 to 1.25 \\ 
 Star & 0.48 & 35 & 1.98 & 0.75 to 1.15 \\
 \hline
 \end{tabular}
 \caption{Comparison of s value}\label{table:t1}
\end{table}

\section{Conclusions} 
\label{cn}
A criterion for selecting the value of Gaussian kernel bandwidth parameter $s$ is proposed in this paper. Good quality data boundary that closely follows data shape can be obtained at values of s where the second derivative of optimal dual objective function value with respect to $s$ first reaches zero. For certain data sets, the method provides a  range of values where this criterion holds good. Any value of s within this range provides a good data boundary. Starting with a very low value of $s$, the search for a good value of $s$ can be abandoned once the second derivative of the optimal objective function reaches zero. As outlined in Section~\ref{rw}, the proposed method provides better results compared to existing methods. The criterion also provides good results when used for high dimensional data. 

\clearpage
\bibliographystyle{plain}
\bibliography{svdd_bw}
\end{document}